\documentclass{article}
\usepackage{iclr2027_conference,times}

\usepackage{amsmath,amsfonts,bm}

\def\eqref#1{equation~\ref{#1}}

\def\1{\bm{1}}

\DeclareMathAlphabet{\mathsfit}{\encodingdefault}{\sfdefault}{m}{sl}
\SetMathAlphabet{\mathsfit}{bold}{\encodingdefault}{\sfdefault}{bx}{n}

\usepackage{amssymb,mathtools}
\usepackage{graphicx}
\usepackage{xcolor}
\usepackage{multirow}
\usepackage{enumitem}
\usepackage{longtable}
\usepackage{makecell}
\usepackage{booktabs}
\usepackage[hidelinks]{hyperref}
\usepackage{url}
\usepackage[normalem]{ulem}
\usepackage{capt-of}
\usepackage[most]{tcolorbox}
\usepackage{wrapfig}

\title{
	\textbf{Highlight-Then-Summarize: Learning to Compress Evidence for Long-Context Understanding}}
\author{
\textbf{Zhaoyuan Xia}$^{1,2,*}$,
\textbf{Qinghongbing Xie}$^{3,*}$,
\textbf{Yung Xiang Hue}$^{3}$,
\textbf{Jianguang Jiang}$^{2}$,
\textbf{Gaofeng Lu}$^{2}$, \\[1mm]
\textbf{Zhenyu Jiao}$^{2}$,
\textbf{Xing Yuan}$^{2}$,
\textbf{Dai Dai}$^{2,\dagger}$,
\textbf{Tong Mo}$^{1,\dagger}$,
\textbf{Long Zeng}$^{3,\dagger}$ \\[2mm]
$^1$Peking University \quad $^2$Baidu Inc. \quad $^3$Tsinghua University \\[1mm]
\texttt{2401210464@stu.pku.edu.cn} \quad
\texttt{xqhb23@mails.tsinghua.edu.cn} \\[1mm]
\texttt{daidai@baidu.com} \quad
\texttt{motong@ss.pku.edu.cn} \quad
\texttt{zenglong@sz.tsinghua.edu.cn}
}

\begin{document}
\iclrfinalcopy
\maketitle
\lhead{Preprint}
\begingroup
\renewcommand{\thefootnote}{\fnsymbol{footnote}}
\footnotetext[1]{Equal contribution.}
\footnotetext[2]{Corresponding authors.}
\endgroup

\begin{abstract}
Long-context understanding is a fundamental capability for large language models to reason over lengthy documents, multi-turn conversations, and code.
However, large language models often struggle with irrelevant and redundant information in long contexts, where task-relevant evidence can be sparse and scattered across distant positions, limiting their ability to reliably solve long context tasks.
To address this problem, we propose \textbf{Highlight-Then-Summarize (H2S)}, a compress-then-reason paradigm that identifies question-relevant evidence from long contexts and summarizes it into a focused representation, enabling LLMs to reason with less interference from irrelevant and redundant information.
We construct H2S-Dataset, comprising 6,647 examples from 11 benchmark families with an average context length of 43.9K tokens, and introduce H2S-Bench for evaluation. H2S-14B, trained on H2S-Dataset, achieves an average score of 32.60 on H2S-Bench, outperforming Qwen3.8-27B by 10.17 points and achieving leading performance among open-source models. Further analysis shows that H2S-14B achieves the highest Evidence-Summary Quality (ESQ) score and retains 97.1\% of its 16K-budget performance with only a 4K output budget, indicating that H2S produces higher-quality yet more compact reasoning.
Our contributions are threefold: 
(1) a compress-then-reason paradigm, H2S, that focuses reasoning on relevant evidence while suppressing irrelevant context; (2) H2S-Dataset, a diverse training set comprising 6,647 examples from 11 benchmark families to support learning under the H2S paradigm; and 
(3) a reinforcement learning method, H2S-RL, with process-level rewards for evidence selection and summary construction.
We hope this work provides a practical approach to improving long-context reasoning through evidence-focused summary. 
Code and H2S-Dataset will be released at \url{https://github.com/X-Luffy/Highlight-Then-Summarize}.

\end{abstract}

\section{Introduction}
\label{sec:background}

Long-context understanding is increasingly important for tasks such as long-document question answering, multi-document reasoning, extended dialogue modeling, and repository-level code understanding\cite{bai2024longbenchv2,jimenez2024swe}.
Despite substantial advances in the context-window capacity of state-of-the-art LLMs~\cite{guo2025deepseek,team2026kimi,openai2025o3}, effective reasoning over long inputs remains difficult because task-relevant evidence is typically sparse, distributed across distant regions of the context, and embedded within substantially more non-essential content~\cite{liu2024lost,hsieh2024ruler}. Beyond evidence grounding, successful reasoning further requires establishing query-conditioned relations among dispersed evidence, including entity associations, temporal dependencies, logical conditions, and numerical constraints. For example, resolving a repository-level bug may require tracing a failing API call across its interface definition, an implementation in another file, and a configuration change introduced elsewhere in the codebase.

\begin{figure*}[htbp]
\centering
\includegraphics[width=\textwidth,trim=10 50 0 10,clip]{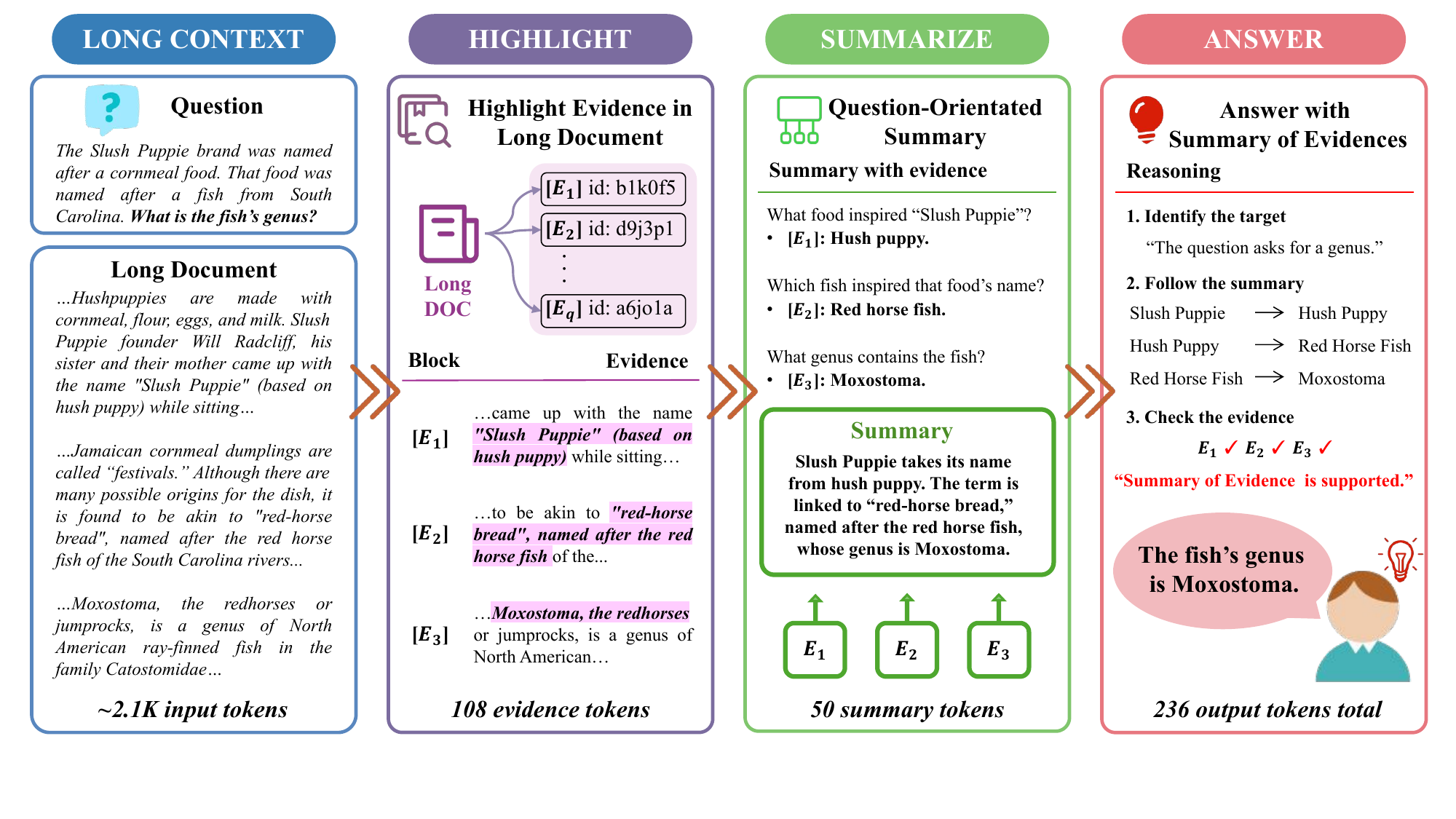}
\caption{\textbf{Highlight-Then-Summarize.} H2S transforms a block-structured long document into source-addressable evidence and then a question-conditioned summary before producing the final answer in one autoregressive trajectory.}
\label{fig:overview}
\end{figure*}

Existing methods primarily improve how models access or select relevant context. Recent long-context reinforcement learning approaches induce structured retrieval-and-reasoning trajectories, directly reward contextual grounding, or optimize heterogeneous long-context capabilities \citep{wang2025loongrl,chen2026longrlvr,lv2026golongrl}. These advances make evidence easier to find, but evidence selection alone does not specify how multiple passages should be combined. Their integration often remains implicit in final-answer generation, where the model must simultaneously retain fragmented evidence, infer the relations among it, and complete the task. This entanglement makes evidence difficult to use consistently and can encourage unnecessarily long reasoning.

We propose \textbf{Highlight-Then-Summarize (H2S)}, a compress-then-reason paradigm that externalizes both evidence selection and evidence integration before final answering, as Figure \ref{fig:overview}. H2S first highlights source-addressable evidence, then organizes the selected information into a question-conditioned summary, and finally produces the answer conditioned on these intermediate representations. 
Highlight preserves a traceable path to the document, whereas Summarize converts fragmented passages into a compact, reasoning-ready state. Both are generated by one autoregressive model, and the original document remains available throughout generation; H2S therefore organizes rather than prunes the context. A motivating study supports this design: adding only 1.04K tokens of self-generated question-conditioned summary to a 15.19K-token document improves Qwen3.5-Flash by 13.3 points.

To supervise this behavior, we construct \textbf{H2S-Dataset}, comprising 6,647 examples from 11 benchmark families, including 4,228 examples for supervised fine-tuning and 2,419 for reinforcement learning. Each example follows the structured Evidence--Summary--Answer format. We further introduce \textbf{H2S-RL}, which uses a deterministic trajectory-level reward to jointly evaluate the final answer, evidence grounding and span coverage, summary quality, and output structure. This design provides direct feedback on both intermediate stages instead of relying only on final-answer correctness. We evaluate the resulting models on \textbf{H2S-Bench}, a seven-task suite spanning retrieval, reasoning, aggregation, citation, and summarization over long contexts.

Under a shared 128K input and 4K output budget, H2S-7B and H2S-14B achieve average scores of 28.56 and 32.60 on H2S-Bench. H2S-14B obtains the strongest overall result among the evaluated open-source models, outperforming the larger QwenLong-L1-32B and Qwen3.8-27B by 5.01 and 10.17 points, respectively. H2S-RL further improves the 7B and 14B supervised checkpoints by 4.88 and 3.44 points. Process-level analyses support the intended mechanism: H2S-14B achieves the highest Evidence--Summary Quality score, removing its summary lowers average performance by 8.47 points, and extending its output budget from 4K to 16K yields only 0.96 additional points. These results indicate that explicitly organizing evidence can improve long-context reasoning without relying on increasingly long generation.

Our contributions are threefold:
\begin{itemize}[leftmargin=*]
    \item[(1)] We propose Highlight-Then-Summarize, a compress-then-reason paradigm that externalizes evidence localization and integration as source-grounded evidence and a question-conditioned summary before final answering.
    \item[(2)] We construct H2S-Dataset, a diverse long-context training set of 6,647 structured Evidence--Summary--Answer examples from 11 benchmark families.
    \item[(3)] We introduce H2S-RL, which jointly evaluates final-answer correctness, evidence grounding and span coverage, summary quality, and output structure. H2S-14B achieves the strongest overall result among the evaluated open-source models on H2S-Bench.
\end{itemize}

\section{Related Work}
\label{sec:related}
\subsection{Long-Context Tasks}

Long-context understanding encompasses a broad set of capabilities rather than a single task. 
Existing benchmarks therefore evaluate complementary aspects of long-context processing. 
LongBench and LongBench~v2 cover a diverse range of tasks, including question answering, summarization, long-form dialogue, code understanding, and reasoning \citep{zhang2024longbench,bai2024longbenchv2}. 
FRAMES emphasizes factual retrieval, multi-document information gathering, and reasoning \citep{krishna2024frames}, while AA-LCR focuses on information synthesis across long documents \citep{artificialanalysis2025aalcr}. 
MRCR evaluates whether models can identify and track requested information in long multi-turn contexts containing distractors \citep{vodrahalli2024michelangelo}, and HELMET assesses retrieval, question answering, summarization, citation, and ranking in application-oriented settings \citep{yen2024helmet}. 

Collectively, these benchmarks assess long-context capabilities spanning information retrieval, cross-context understanding, and reasoning over distributed evidence. These capabilities require models to move beyond locating relevant information toward selecting and organizing evidence across the context. However, existing benchmarks focus primarily on final outputs, offering limited assessment of this intermediate evidence-processing stage. Our work targets this gap by measuring how models identify, consolidate, and summarize distributed evidence before reasoning.

\subsection{Evidence Processing for Long-Context Reasoning}

\textbf{{Evidence Retrieval and Attribution.}}
Retrieval-augmented generation decouples information access from answer generation by retrieving relevant passages as external evidence \citep{lewis2020retrieval}. Lexical retrieval, dense retrieval, and rank fusion constitute standard approaches for evidence localization \citep{robertson2009probabilistic,cormack2009reciprocal}. 
Attribution-oriented methods instead emphasize the correspondence between generated claims and supporting sources: ALCE evaluates citation correctness and completeness \citep{gao2023alce}, Self-RAG integrates retrieval with self-reflective generation \citep{asai2023selfrag}, and LongCite introduces fine-grained citation supervision for long-context question answering \citep{zhang2024longcite}. 
Collectively, these methods concern the localization of relevant evidence and its attribution to source text. H2S introduces explicit supervision for evidence identification, positioning question-relevant evidence extraction as a distinct intermediate objective preceding summarization and reasoning.

\textbf{{Context Engineering for Compression.}}
Question-aware prompt compression selectively preserves relevant information in long inputs \citep{jiang2024longllmlingua}. Recent work also compresses growing interaction histories into compact states for continued reasoning. ReSum periodically summarizes agent trajectories into concise reasoning states for long-horizon search \citep{wu2025resum}. A-MEM constructs and updates structured memory notes with explicit links among stored information \citep{xu2025amem}, while Memory-R1 learns memory operations for retaining, updating, and discarding information \citep{yan2025memoryr1}. 
These methods perform context compression as an external mechanism for managing long interaction histories or persistent memories. In contrast, H2S internalizes compression into the reasoning process by supervising the construction of a question-conditioned evidence summary as an intermediate representation for downstream reasoning.

\subsection{Long-Context Reinforcement Learning}

Recent work has explored reinforcement learning for improving information acquisition and utilization in long-context settings. 
LoongRL constructs distractor-rich, multi-step tasks to train models for planning, retrieval, reasoning, and verification \citep{wang2025loongrl}. 
LongRLVR introduces verifiable context rewards that supervise the context selected prior to answer generation \citep{chen2026longrlvr}. 
GoLongRL extends long-context reinforcement learning to multiple capability-oriented tasks and employs reward balancing for multitask alignment \citep{lv2026golongrl}. Existing approaches mainly optimize evidence retrieval or context selection, while leaving the subsequent organization of multiple evidence items unsupervised. Our method instead applies process supervision to both evidence selection and question-conditioned summary construction, explicitly optimizing the transition from retrieved evidence to a compact intermediate representation before reasoning.

\section{METHODOLOGY}
\label{sec:method}
\subsection{Why a Few Summary Tokens Help}

\begin{wraptable}{r}{0.43\textwidth}
\vspace{-25pt}
\centering
\scriptsize
\setlength{\tabcolsep}{2.5pt}
\renewcommand{\arraystretch}{1.05}

\caption{\textbf{Even a few summary tokens can substantially improve long-document QA.}}
\label{tab:summary_motivation}

\resizebox{\linewidth}{!}{%
\begin{tabular}{lrr}
\toprule
Input to Qwen3.5-Flash & Context tokens & Acc. \\
\midrule
LongDoc & 15.19K & 72.6 \\
Claude Summary & 0.46K & 93.0 \\
LongDoc + Claude Summary & 15.19K + 0.46K & \textbf{96.7} \\
LongDoc + Self-Summary (S) & 15.19K + 0.14K & 69.3 \\
LongDoc + Self-Summary (M) & 15.19K + 0.60K & 77.8 \\
LongDoc + Self-Summary (L) & 15.19K + 1.04K & 85.9 \\
\bottomrule
\end{tabular}%
}
\vspace{-8pt}
\end{wraptable}

Cognitive theories of text comprehension suggest that effective understanding relies on organizing local information into a goal-conditioned macrostructure~\cite{kintsch1978toward}. This motivates a natural question: \textbf{Can a compact, question-conditioned representation improve reasoning over long documents?} To investigate this question, we evaluate question-conditioned summaries on DocQA-RL-1.6K~\citep{wan2025qwenlongl1}, using Qwen3.5-Flash as the answer model. We compare reasoning over the original document with reasoning over summaries generated solely from the question and document, without access to the reference answer. \textit{Claude Summary} is generated by Claude Sonnet 4.6, whereas \textit{Self-Summary} is produced by Qwen3.5-Flash in a separate pass.

As shown in Table~\ref{tab:summary_motivation}, a compact question-conditioned summary can substantially improve downstream reasoning despite containing only a small fraction of the original context. A 0.46K-token Claude summary, approximately 3\% of the 15.19K-token document, increases accuracy from 72.6 to 93.0 when used alone, while combining it with the original document further improves performance to 96.7. Similar gains are observed with self-generated summaries: a 1.04K-token self-summary improves accuracy from 72.6 to 85.9. These results suggest that explicitly restructuring distributed evidence into a compact, question-conditioned representation can make long-context information more effective for subsequent reasoning, providing direct motivation for the evidence-to-summary stage in H2S. Full results across answer models and document-length ranges are reported in Appendix~\ref{sec:appendix_summary_motivation}.

\subsection{Highlight-Then-Summarize Paradigm}

\textbf{Highlight-Then-Summarize (H2S)} is a compress-then-reason paradigm that externalizes evidence selection and evidence integration as explicit intermediate representations before final-answer generation. Given a question $q$ and a long document divided into source-addressable blocks $D=(B_1,\ldots,B_M)$, H2S follows
\begin{equation}
(D,q)
\xrightarrow{\text{Highlight}} E_q
\xrightarrow{\text{Summarize}} S_q
\xrightarrow{\text{Reason}} a.
\label{eq:h2s_pipeline}
\end{equation}
The Highlight stage produces question-relevant evidence
\begin{equation}
E_q=\{(u_i,s_i)\}_{i=1}^{m}, \qquad s_i \subseteq B(u_i),
\end{equation}
where $u_i$ is a stable source-block identifier, $B(u_i)$ is the corresponding block, and $s_i$ is its supporting span. The Summarize stage then maps these traceable but potentially scattered spans into a question-conditioned summary $S_q$ that retains their answer-relevant content and makes their relations explicit. Thus, $S_q$ is neither a generic document summary nor a concatenation of retrieved passages, but a compact representation organized for the reasoning required by $q$.

H2S realizes the three operations within a single autoregressive trajectory. Its conditional distribution factorizes as
\begin{equation}
p_\theta(E_q,S_q,a\mid D,q)
=p_\theta(E_q\mid D,q)\,
p_\theta(S_q\mid D,q,E_q)\,
p_\theta(a\mid D,q,E_q,S_q).
\label{eq:h2s_factorization}
\end{equation}
The three factors respectively model source-grounded localization, question-conditioned evidence integration, and task completion. Unlike direct supervision of $p_\theta(a\mid D,q)$, this factorization makes the first two operations observable and therefore available for supervision and process-level evaluation.

As shown in Figure~\ref{fig:overview}, Highlight and Summarize form complementary interfaces: $E_q$ preserves provenance but may remain fragmented, whereas $S_q$ integrates the selected information into a compact reasoning state. Both are generated by the same model, and the original document remains available throughout generation. H2S therefore organizes the context rather than pruning it, preserving access to the source while making the information needed for reasoning explicit.

\subsection{Construction of H2S-Dataset}

\begin{figure*}[t]
\centering

\begin{minipage}[t]{0.62\textwidth}
    \centering
    \includegraphics[
        width=0.9\textwidth,
        trim=0 0 0 40,
        clip
    ]{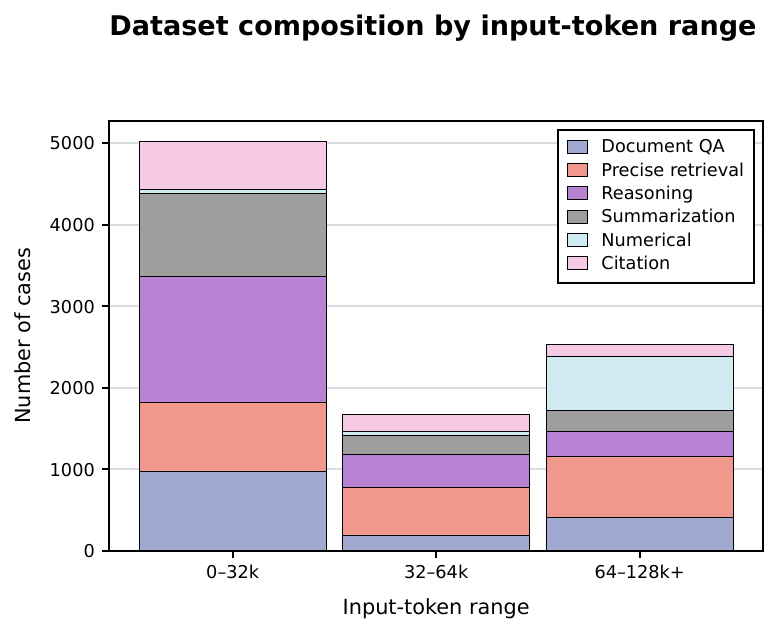}

    \vspace{-2mm}
    \textbf{(a) Length Analysis of H2S Dataset}
\end{minipage}
\hfill
\begin{minipage}[t]{0.37\textwidth}
    \centering
    \includegraphics[
        width=\textwidth,
        trim=0 0 0 0,
        clip
    ]{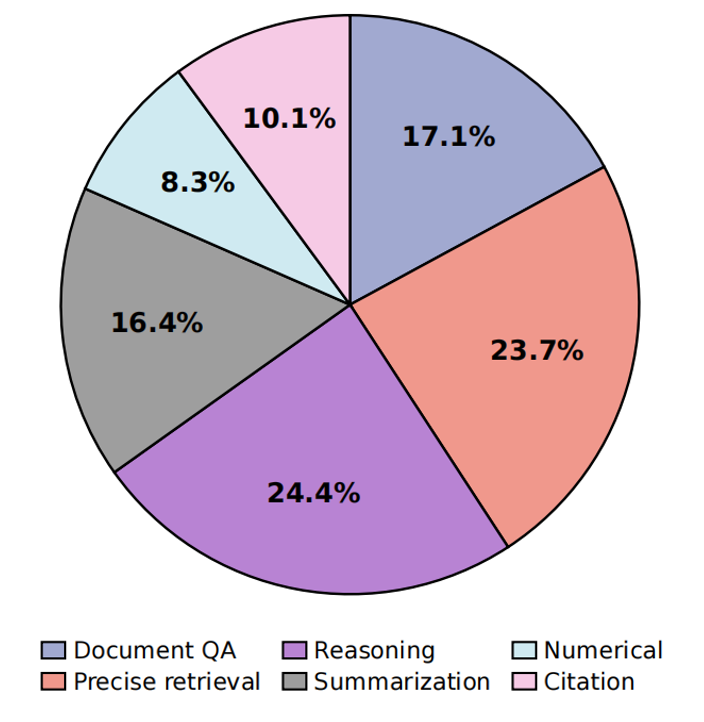}

    \vspace{-2mm}
    \textbf{(b) Task Category Analysis}
\end{minipage}

\vspace{-1mm}
\caption{
    \textbf{Composition of H2S-Dataset.}
    H2S-Dataset spans diverse long-context tasks and input lengths, with reasoning and precise retrieval accounting for 24.4\% and 23.7\% of the dataset, respectively, while examples cover input lengths from 0--32K to 64--128K+ tokens.
}
\label{fig:dataset_composition}
\vspace{-3mm}
\end{figure*}






Most long-context datasets supervise only the final answer, leaving $E_q$ and $S_q$ in Equation~\ref{eq:h2s_factorization} unobserved. Following prior work on long-context alignment data \citep{bai2024longalign}, we construct H2S-Dataset by augmenting existing instances $(D,q,a)$ with source-grounded evidence and a question-conditioned summary, yielding $(D,q,E_q,S_q,a)$. H2S-Dataset comprises 6,647 examples from 11 benchmark families, with 4,228 examples for SFT and 2,419 for RL. The construction pipeline shows as Figure \ref{fig:data_construction}. Source and length distributions are reported in Appendix~\ref{sec:appendix_data}.

\begin{figure*}[htbp]
\centering
\includegraphics[width=\textwidth,trim=0 35 0 10, clip]{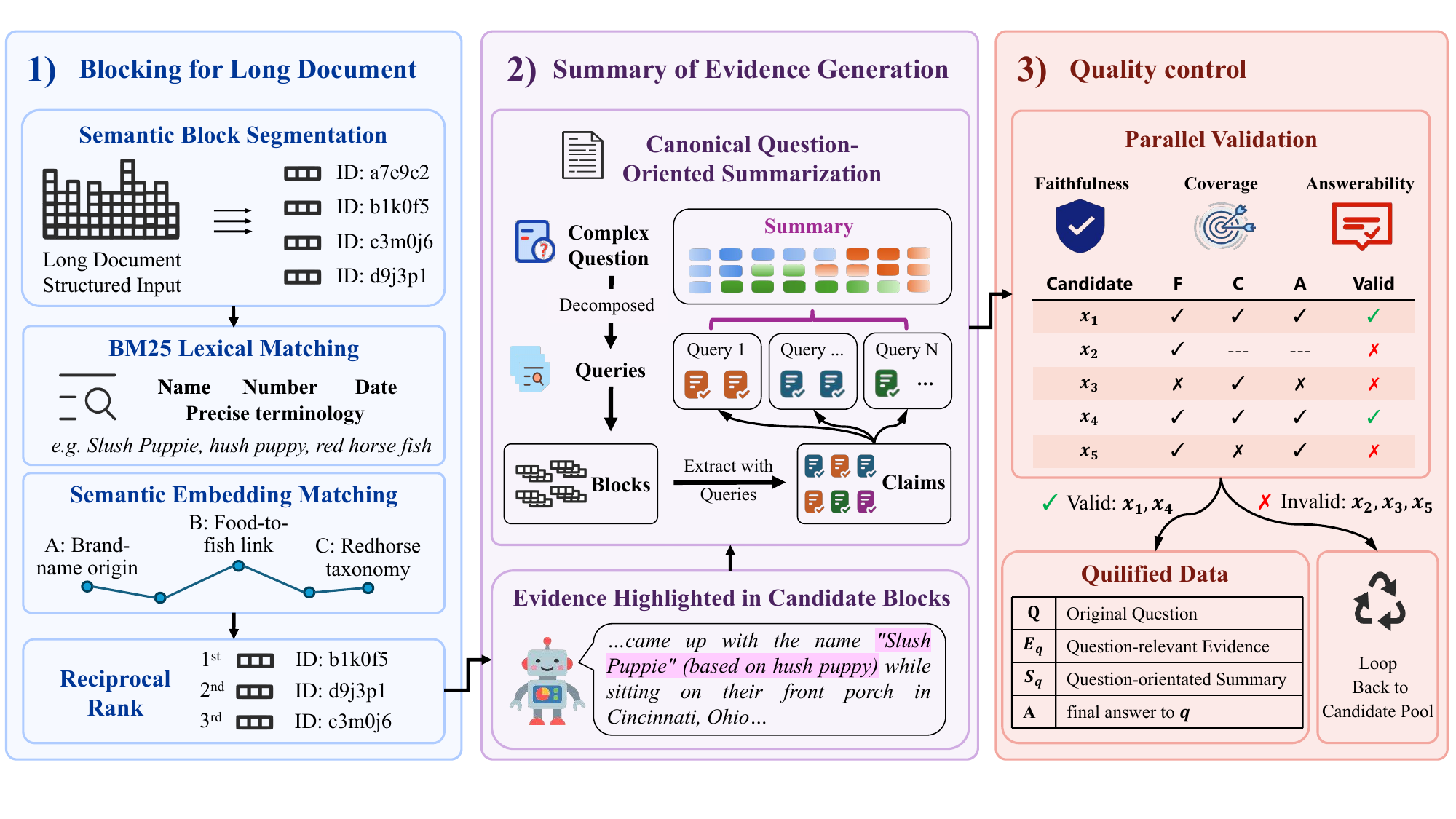}
\caption{\textbf{Dataset Construction of Highlight-Then-Summarize.} Each training instance is constructed through semantic blocking, candidate retrieval, query-guided evidence extraction, question-conditioned summarization, and quality verification, with failed instances iteratively reconstructed until the required criteria are satisfied.}
\label{fig:data_construction}
\end{figure*}

\textbf{{Blocking for Long Document.}}
We first segment each document at semantic boundaries and assign every block a stable identifier while preserving its source position. Hybrid lexical and semantic retrieval forms a high-recall candidate pool. We then decompose the question into atomic information needs and extract question-relevant claims, binding each claim to a \texttt{BLOCK\_ID} and its supporting span. This produces traceable evidence supervision $E_q$. Retrieval is used only for offline data construction; H2S reads the full supplied document without an external retriever at inference time.

\textbf{{Summary of Evidence Generation.}}
We remove unsupported and duplicate claims, organize the retained claims by information need, and synthesize the question-conditioned summary $S_q$. The information needs act as a coverage scaffold, preventing a fluent summary from omitting part of the question. Consequently, $S_q$ integrates source-linked claims rather than concatenating retrieved passages. A complete construction example is provided in Appendix~\ref{sec:appendix_construction_case}.

\textbf{{Quality control.}}
We retain an instance only if it passes three checks: \emph{faithfulness}, which verifies block identifiers and source-span alignment; \emph{coverage}, which requires every information need to be supported; and \emph{answerability}, which tests whether the answer can be derived from the question and constructed summary. Failed instances return for reconstruction. The accepted record preserves the original task answer and adds $E_q$ and $S_q$ as supervision for evidence localization and information integration.

\subsection{H2S-RL}

\textbf{H2S-RL} employs GRPO \citep{shao2024deepseekmath} to optimize each generated trajectory $y=(E,S,a)$ with a deterministic programmatic reward that jointly evaluates the final answer and the evidence-to-summary process, without an online LLM judge. It uses three quality signals and one structure gate. \textbf{(1) Answer quality} $R_{\mathrm{answer}}$ follows the native output contract of each task. \textbf{(2) Evidence quality} $R_{\mathrm{evidence}}$ comprises two complementary components: $R_{\mathrm{ground}}$ is the fraction of evidence items whose block identifiers and span locators can be resolved in the source document, while $R_{\mathrm{span}}$ is the span-level F1 between the resolved and reference evidence intervals. \textbf{(3) Summary quality} $R_{\mathrm{summary}}$ is ROUGE-L \citep{lin2004rouge} between the generated and reference question-conditioned summaries after removing local evidence identifiers. \textbf{(4) Structure quality} $R_{\mathrm{structure}}$ gates trajectories with missing, partially valid, or complete Evidence--Summary--Answer structure and source fields.

\begin{equation}
\small
\mathcal{R}_{\mathrm{H2S}}(y)=R_{\mathrm{structure}}(y)\left[\lambda R_{\mathrm{answer}}(a)+(1-\lambda)\frac{2R_{\mathrm{evidence}}(E)R_{\mathrm{summary}}(S)}{R_{\mathrm{evidence}}(E)+R_{\mathrm{summary}}(S)}\right].
\label{eq:h2s_reward}
\end{equation}
This design treats evidence selection and summary construction as complementary stages: the process reward remains low when either stage fails, rather than allowing one strong component to conceal the other. The structure gate further prevents malformed trajectories from exploiting otherwise high component scores. Consequently, the within-prompt advantages used by GRPO favor outputs that improve task success through a complete and source-grounded intermediate process, without requiring an online LLM judge. Metric routing and optimization details are reported in Appendix~\ref{sec:appendix_setup}.

\section{Experiments}
\label{sec:experiments}

\subsection{Experimental Setup}


We train H2S-7B and H2S-14B initialized from Qwen2.5-7B/14B-Instruct, using full-parameter supervised SFT followed by GRPO. Training adopts a progressive context-length curriculum of 32K$\rightarrow$64K$\rightarrow$128K. We evaluate the resulting models from a suite of 2,575 examples spanning seven long-context capabilities: numerical reasoning (DocFinQA), multi-document reasoning (Frames), citation-grounded answering (LongCite), precise retrieval (MRCR), distributed-information aggregation (AA-LCR), general long-context understanding (LongBenchV2), and long-document summarization (HELMET-Summ).
Performance is assessed exclusively on the final answer using the native evaluation metric of each benchmark.
All models are evaluated under an identical input configuration, receiving the same system prompt, question, and block-structured document, with a maximum input context of 128K tokens and an output budget of 4K tokens. 
Further details on the training and evaluation protocol are provided in Appendix~\ref{sec:appendix_setup} and Appendix~\ref{sec:appendix_curriculum}.

\subsection{Main Results}
\begin{table*}[htbp]
\centering
\scriptsize
\setlength{\tabcolsep}{2.8pt}
\renewcommand{\arraystretch}{1.08}
\caption{\textbf{Main results on seven long-context benchmarks.}
All models use a maximum input length of 128K and a maximum output length of 4K. Closed-source models are included for reference; among all open-source models, including our models, the best result in each column is shown in \textbf{bold}, and the second-best result is shown in \underline{\textbf{bold and underlined}}.}
\label{tab:main_results}
\resizebox{\textwidth}{!}{%
\begin{tabular}{llcccccccc}
\toprule
Category & Model & DocFinQA & Frames & LongCite & MRCR & AA-LCR & LongBenchV2 & HELMET-Summ & Average \\
\midrule
\multirow{5}{*}{Closed-source}
& Claude Opus 5 & 56.11 & 27.89 & 12.17 & 100.00 & 26.45 & 64.18 & 15.13 & 43.13 \\
& GPT-5.5 & 49.57 & 36.05 & 12.62 & 92.38 & 52.34 & 46.15 & 9.45 & 42.65 \\
& DeepSeek-V4-Pro & 34.83 & 37.41 & 6.78 & 72.14 & 47.77 & 48.56 & 12.48 & 37.14 \\
& Kimi-K2.5 & 51.14 & 37.41 & 10.83 & 87.58 & 39.95 & 50.00 & 14.41 & 41.62 \\
& GLM-5.2 & 49.57 & 36.05 & 12.62 & 92.38 & 40.85 & 48.80 & 16.52 & 42.40 \\
\midrule
\multirow{9}{*}{Open-source}
& DeepSeek-R1-Distill-Qwen-7B & 0.06 & 8.50 & 1.31 & 0.40 & 0.06 & 1.92 & 5.29 & 2.51 \\
& DeepSeek-R1-Distill-Qwen-14B & 11.69 & 23.13 & 5.25 & 38.28 & 8.75 & 23.32 & 11.81 & 17.46 \\
& Qwen3-4B-Thinking-2507 & 14.61 & 18.71 & 3.68 & 25.45 & 9.88 & 18.99 & 10.80 & 14.59 \\
& Qwen3.8-27B & \textbf{40.52} & 32.31 & 4.94 & 46.89 & 15.69 & 15.14 & 1.51 & 22.43 \\
& Qwen3.5-35B-A3B & 24.72 & 18.03 & 2.13 & 8.62 & 6.11 & 6.97 & 0.06 & 9.52 \\
& QwenLong-L1-32B & \underline{\textbf{35.03}} & 34.35 & 8.16 & 40.28 & \underline{\textbf{26.20}} & \textbf{37.98} & 11.13 & 27.59 \\
& QwenLong-L1.5-30B-A3B & 9.34 & 25.17 & 1.03 & 0.00 & 5.37 & 30.52 & 3.92 & 10.76 \\
& Qwen2.5-7B-Instruct-1M & 5.00 & 13.95 & 5.37 & 4.01 & 7.31 & 24.04 & 7.16 & 9.55 \\
& Qwen2.5-14B-Instruct-1M & 15.70 & 23.47 & 8.32 & 25.05 & 13.56 & \underline{\textbf{35.82}} & \underline{\textbf{12.03}} & 19.14 \\
\midrule
\multirow{4}{*}{\textbf{Ours}}
& H2S-7B-SFT & 20.45 & 31.29 & 8.65 & 58.12 & 13.96 & 27.88 & 5.42 & 23.68 \\
& H2S-7B & 26.33 & 32.97 & \textbf{15.76} & 59.72 & 21.47 & 32.11 & 11.58 & 28.56 \\
\cmidrule(lr){2-10}
& H2S-14B-SFT & 31.49 & \textbf{37.76} & 10.50 & \underline{\textbf{65.13}} & 22.40 & 28.37 & 8.45 & \underline{\textbf{29.16}} \\
& H2S-14B & 30.83 & \underline{\textbf{37.72}} & \underline{\textbf{13.88}} & \textbf{66.13} & \textbf{26.86} & 35.42 & \textbf{17.38} & \textbf{32.60} \\
\bottomrule
\end{tabular}}
\end{table*}


\textbf{H2S demonstrates strong overall performance across long-context benchmarks.}
As shown in Table~\ref{tab:main_results}, H2S-14B achieves the highest average score among the evaluated open-source models, reaching 32.60 and surpassing the larger QwenLong-L1-32B and Qwen3.8-27B by 5.01 and 10.17 points, respectively. H2S-7B attains an average score of 28.56, exceeding QwenLong-L1-32B by 0.97 points despite using fewer than one quarter of its parameters. Moreover, H2S-14B ranks among the top two open-source models on five of the seven benchmarks, spanning cited answering, precise retrieval, distributed-information aggregation, and long-document summarization. These results suggest that the gains of H2S generalize across heterogeneous long-context reasoning settings rather than being driven by performance on a particular benchmark or task type.

\textbf{H2S-RL delivers consistent gains over SFT.}
Starting from the SFT checkpoints, H2S-RL raises the average by 4.88 points for 7B and 3.44 points for 14B. These gains are consistent across model scales and are concentrated on tasks that require selecting and integrating information distributed through long inputs: for example, H2S-RL improves LongCite, AA-LCR, and HELMET-Summ by 7.11, 7.51, and 6.16 points for 7B, and by 3.38, 4.46, and 8.93 points for 14B. This consistent improvement over strong SFT checkpoints shows that trajectory-level optimization adds value beyond learning the output structure by imitation. The following process-level analyses examine whether these final-answer gains are accompanied by better evidence and summaries.

\subsection{Analysis of Summarization Quality}
\label{sec:esq}

To determine whether better final answers arise from a better intermediate process, we introduce \textbf{Evidence--Summary Quality (ESQ)}. 
ESQ is high only when a model both recovers the reference evidence blocks and produces a reference-aligned question-conditioned summary; it is computed as their case-level harmonic mean followed by macro-averaging. 
\begin{wrapfigure}{r}{0.6\textwidth}
\vspace{-10pt}
 \centering
    \includegraphics[
        width=\linewidth,
        trim=2mm 5mm 0mm 2mm,
        clip
    ]{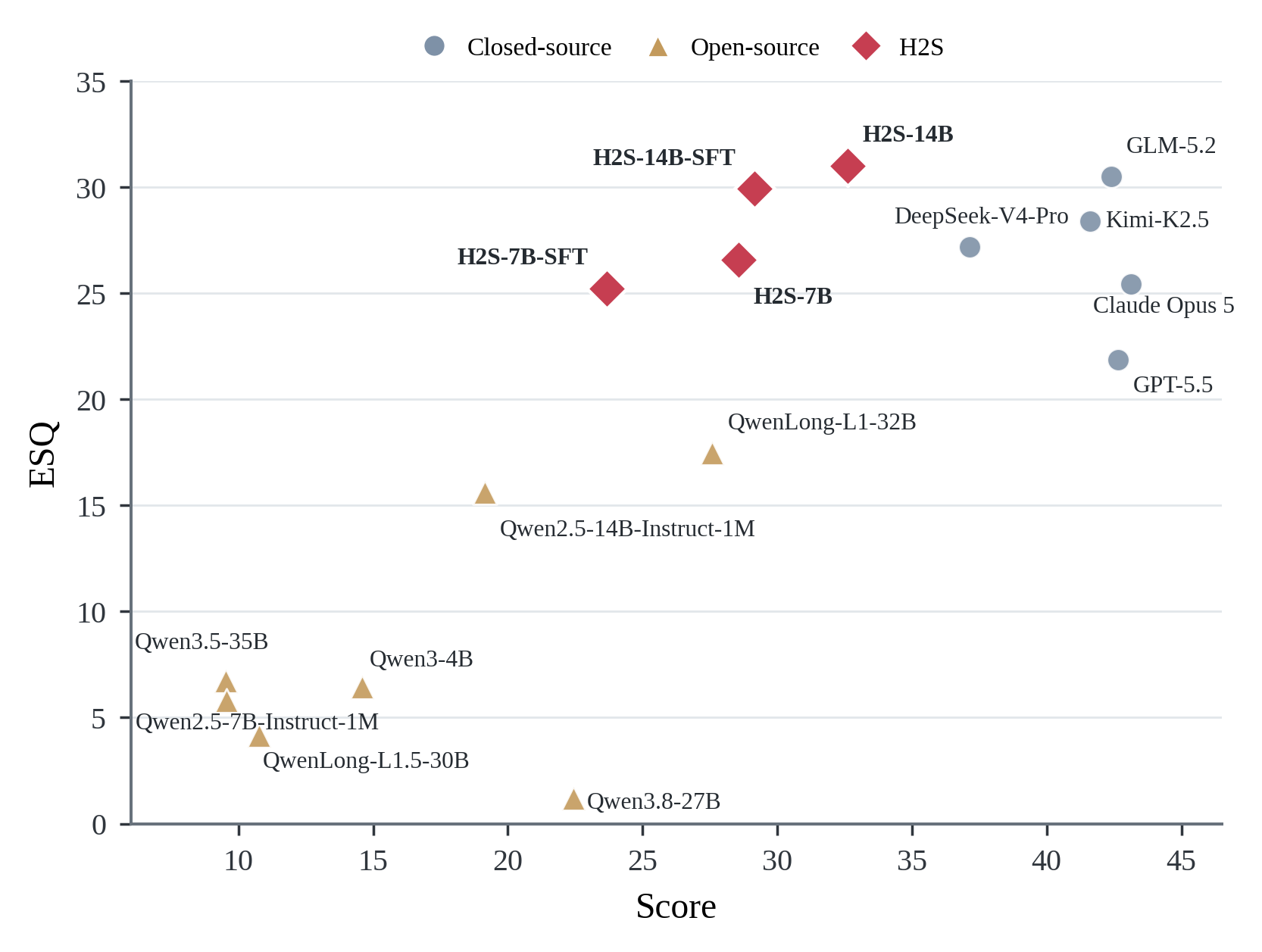}

    \caption{\textbf{Analysis of Summarization Quality.}
    Final-answer score versus Evidence--Summary Quality (ESQ).
    Higher is better on both axes.}
    \label{fig:esq_training_paths}

\vspace{-10pt}
\end{wrapfigure}
The complete definition and per-model component scores are provided in Appendix~\ref{sec:appendix_esq}. 
The upper-right region of Figure~\ref{fig:esq_training_paths} corresponds to strong final answers together with high-quality intermediate representations. H2S-14B reaches the highest ESQ among all evaluated models at 31.02, followed by GLM-5.2 at 30.50; H2S-7B reaches 26.60 and substantially outperforms the open-source baselines. Both H2S-SFT and H2S improve markedly over their corresponding base models, while H2S-RL further raises ESQ from 25.22 to 26.60 for 7B and from 29.93 to 31.02 for 14B. The gains in final-answer score are therefore accompanied by better evidence selection and summary construction rather than answer fitting alone.

\subsection{Ablations study}
\begin{table*}[h]
\centering
\scriptsize
\setlength{\tabcolsep}{3.2pt}
\renewcommand{\arraystretch}{1.08}
\caption{\textbf{Detailed ablation of evidence and summary generation.}
\textit{Baseline} removes both evidence and summary generation;
\textit{w/o evidence} and \textit{w/o summary} remove the corresponding intermediate component, respectively.
Avg is the unweighted macro-average over the seven tasks.}
\label{tab:component_ablation_full}
\resizebox{\textwidth}{!}{%
\begin{tabular}{lrrrrrrrr}
\toprule
Model & DocFinQA & Frames & LongCite & MRCR & AA-LCR & LongBenchV2 & HELMET-Summ & Avg \\
\midrule

Baseline-7B
& 4.82 & 21.43 & 3.92 & 34.80 & 10.71 & 29.09 & 2.65 & 15.35 \\

w/o evidence-7B
& 23.56 & 30.61 & 9.97 & 54.60 & 14.60 & 31.01 & 7.52 & 24.55 \\

w/o summary-7B
& 13.64 & 29.93 & 4.23 & 42.40 & 10.98 & 21.63 & 7.81 & 18.66 \\

\textbf{H2S-7B}
& 26.33 & 32.97 & 15.76 & 59.72 & 21.47 & 32.11 & 11.58 & \textbf{28.56} \\

\midrule

Baseline-14B
& 25.52 & 29.59 & 4.06 & 39.00 & 13.19 & 31.73 & 12.89 & 22.28 \\

w/o evidence-14B
& 25.84 & 37.41 & 10.64 & 59.80 & 17.34 & 35.82 & 11.60 & 28.35 \\

w/o summary-14B
& 24.36 & 33.67 & 4.94 & 47.40 & 17.86 & 28.61 & 12.09 & 24.13 \\

\textbf{H2S-14B}
& 30.83 & 37.72 & 13.88 & 66.13 & 26.86 & 35.42 & 17.38 & \textbf{32.60} \\

\bottomrule
\end{tabular}}
\end{table*}

\textbf{Evidence and Summary Ablation.}  We ablate the two intermediate components at inference time while retaining the final-answer stage. During autoregressive decoding, we mask the ablated component from subsequent attention while keeping the input, checkpoint, and final-answer stage unchanged. Table~\ref{tab:component_ablation_full} reports the average performance under the same 128K input and 4K output limits as the main evaluation. Both components are necessary. Removing evidence lowers the average by 4.01 points for H2S-7B and 4.25 for H2S-14B, showing that explicit localization provides information that cannot be recovered reliably from the raw context during answering. Removing the summary is more damaging, with drops of 9.90 and 8.47 points; removing both causes the largest drops, 13.21 and 10.32 points. These results support our central claim that localization alone is insufficient: scattered evidence must be compressed into a question-conditioned representation before reasoning.

\textbf{Output Budget Ablation.}
Table~\ref{tab:budget} compares H2S with open-source baselines under maximum output budgets of 4K, 8K, and 16K tokens. The input length is fixed at 128K for all settings, and task-level results are reported in Appendix~\ref{sec:appendix_output_budget}. Increasing the output budget substantially improves several thinking baselines. For example, Qwen3.8-27B increases from 22.43 at 4K to 43.72 at 16K. In contrast, H2S is substantially less sensitive to the output budget: H2S-7B improves by only 1.92 points from 4K to 16K, while H2S-14B improves by 0.96 points, with most of the gain already realized at 8K. Notably, H2S-14B reaches 32.60 under the 4K budget, outperforming Qwen3.8-27B by 10.17 points at the same output length. These results indicate that H2S achieves strong long-context performance with a relatively modest generation budget.

\subsection{Attention Distribution of Long Document Analysis}
\label{sec:attention_case_study}

\begin{figure*}[t]
\vspace{-6pt}
\centering

\begin{minipage}[t]{0.54\textwidth}
\vspace{0pt}
\centering
\includegraphics[
    width=\linewidth,
    height=4.6cm,
    keepaspectratio,
    trim=2mm 2mm 2mm 2mm,
    clip
]{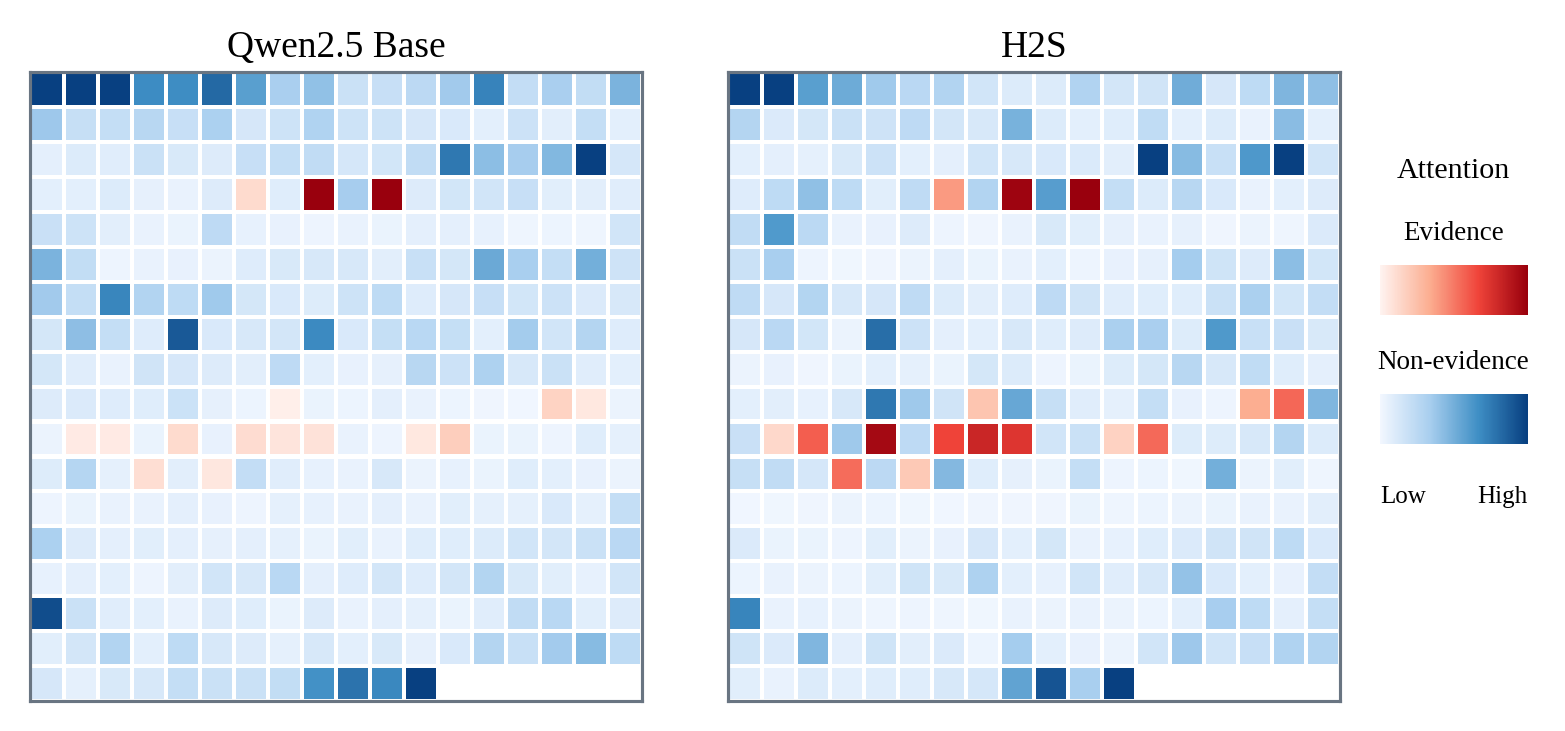}

\caption{\textbf{Attention Distribution of Long Document Analysis.}
H2S improves long-context localization by inducing more concentrated attention over query-relevant evidence in the source document.}
\label{fig:attention_case_study}
\end{minipage}
\hfill
\begin{minipage}[t]{0.44\textwidth}
\vspace{0pt}
\centering
\scriptsize
\setlength{\tabcolsep}{2.8pt}
\renewcommand{\arraystretch}{1.05}

\resizebox{\linewidth}{!}{%
\begin{tabular}{lccc}
\toprule
\multirow{2}{*}{\textbf{Model}}
& \multicolumn{3}{c}{\textbf{Score}} \\
\cmidrule(lr){2-4}
& 4K & 8K & 16K \\
\midrule
Qwen3-4B-Thinking-2507  & 14.59 & 22.84 & 26.21 \\
Qwen3.8-27B             & 22.43 & 37.56 & 43.72 \\
Qwen3.5-35B-A3B         & 9.52  & 30.44 & 35.15 \\
QwenLong-L1.5-30B-A3B   & 10.76 & 18.62 & 25.56 \\
H2S-7B                   & 28.56 & 30.39 & 30.48 \\
H2S-14B                  & 32.60 & 33.44 & 33.56 \\
\bottomrule
\end{tabular}%
}

\caption{\textbf{Output budget ablation.}
Average score under different output budgets.}
\label{tab:budget}
\end{minipage}

\vspace{-8pt}
\end{figure*}

To inspect how H2S changes document reading, we compare block-level attention from Qwen2.5-7B-Instruct-1M and H2S-7B at the onset of evidence generation. Figure~\ref{fig:attention_case_study} shows all 318 source blocks from the same tokenized LongCite input in document order. H2S more than doubles the share of attention assigned to the 16 annotated evidence blocks while leaving attention over most non-evidence blocks comparatively diffuse. This source-traceable case study illustrates the evidence-focused reading behavior that complements the aggregate ablation results; it is a mechanism example rather than a performance estimate.

\section{Conclusion}

We present Highlight-Then-Summarize (H2S), a compress-then-reason paradigm that explicitly structures long-context reasoning through source-grounded evidence localization and question-conditioned summarization. Supported by H2S-Dataset and H2S-RL, H2S consistently improves final-answer performance and intermediate reasoning quality across seven long-context benchmarks while retaining strong performance under limited output budgets. A current limitation is that H2S is less naturally suited to tasks requiring broadly distributed context coverage, while process-level evaluation remains dependent on construction-time evidence and summary references. Future work will explore more adaptive compression objectives and stronger process supervision, with the broader goal of making long-context reasoning more structured, efficient, and reliable.

\section*{REPRODUCIBILITY STATEMENT}
We support reproducibility through detailed descriptions of the method, data, training, and evaluation protocols, together with a public code release. Section~\ref{sec:method} specifies the H2S factorization, data-construction pipeline, and H2S-RL reward. Section~\ref{sec:experiments} defines the shared evaluation setting and reports the main comparisons and ablations. Appendix~\ref{sec:appendix_setup} provides model initialization, optimization hyperparameters, and evaluation details; Appendices~\ref{sec:appendix_data} and~\ref{sec:appendix_curriculum} report the training-data statistics and the complete context-length curriculum. Appendix~\ref{sec:appendix_esq} defines ESQ and reports its component scores, while Appendix~\ref{sec:appendix_output_budget} provides task-level results under different output budgets. We additionally provide a complete data-construction example in Appendix~\ref{sec:appendix_construction_case} and qualitative success and failure cases in Appendix~\ref{sec:appendix_detailed_cases}. The public repository includes data-processing utilities, H2S-SFT and H2S-RL configurations, the programmatic reward, task-specific evaluation code, data schemas, and executable examples. Code and H2S-Dataset will be released publicly upon publication.

\bibliographystyle{iclr2027_conference}
\bibliography{iclr2027_conference}

@article{bai2024longbenchv2,
  title={LongBench v2: Towards Deeper Understanding and Reasoning on Realistic Long-context Multitasks},
  author={Bai, Yushi and others},
  journal={arXiv preprint arXiv:2412.15204},
  year={2024}
}

@article{krishna2024frames,
  title={Fact, Fetch, and Reason: A Unified Evaluation of Retrieval-Augmented Generation},
  author={Krishna, Satyapriya and others},
  journal={arXiv preprint arXiv:2409.12941},
  year={2024}
}

@misc{artificialanalysis2025aalcr,
  title={Artificial Analysis Long Context Reasoning Benchmark},
  author={{Artificial Analysis}},
  year={2025},
  howpublished={Benchmark}
}

@article{vodrahalli2024michelangelo,
  title={Michelangelo: Long Context Evaluations Beyond Haystacks via Latent Structure Queries},
  author={Vodrahalli, Kiran and others},
  journal={arXiv preprint arXiv:2409.12640},
  year={2024}
}

@article{yen2024helmet,
  title={HELMET: How to Evaluate Long-Context Language Models Effectively and Thoroughly},
  author={Yen, Howard and others},
  journal={arXiv preprint arXiv:2410.02694},
  year={2024}
}

@article{gao2023alce,
  title={Enabling Large Language Models to Generate Text with Citations},
  author={Gao, Tianyu and Yen, Howard and Yu, Jiatong and Chen, Danqi},
  journal={arXiv preprint arXiv:2305.14627},
  year={2023}
}

@article{asai2023selfrag,
  title={Self-RAG: Learning to Retrieve, Generate, and Critique through Self-Reflection},
  author={Asai, Akari and Wu, Zeqiu and Wang, Yizhong and Sil, Avirup and Hajishirzi, Hannaneh},
  journal={arXiv preprint arXiv:2310.11511},
  year={2023}
}

@article{zhang2024longcite,
  title={LongCite: Enabling LLMs to Generate Fine-grained Citations in Long-context QA},
  author={Zhang, Jiajie and Bai, Yushi and Lv, Xin and others},
  journal={arXiv preprint arXiv:2409.02897},
  year={2024}
}

@article{wu2025resum,
  title={ReSum: Improving Long-Horizon Search via Context Summarization},
  author={Wu, Xixi and others},
  journal={arXiv preprint arXiv:2509.13313},
  year={2025}
}

@article{xu2025amem,
  title={A-MEM: Agentic Memory for LLM Agents},
  author={Xu, Wujiang and Liang, Zujie and Mei, Kai and Gao, Hang and Tan, Juntao and Zhang, Yongfeng},
  journal={arXiv preprint arXiv:2502.12110},
  year={2025}
}

@article{yan2025memoryr1,
  title={Memory-R1: Enhancing Large Language Model Agents to Manage and Utilize Memories via Reinforcement Learning},
  author={Yan, Sikuan and Yang, Xiufeng and Huang, Zuchao and Nie, Ercong and Ding, Zifeng and Li, Zonggen and Ma, Xiaowen and Schuetze, Hinrich and Tresp, Volker and Ma, Yunpu},
  journal={arXiv preprint arXiv:2508.19828},
  year={2025}
}

@article{kintsch1978toward,
  title={Toward a Model of Text Comprehension and Production},
  author={Kintsch, Walter and van Dijk, Teun A.},
  journal={Psychological Review},
  volume={85},
  number={5},
  pages={363--394},
  year={1978},
  publisher={American Psychological Association}
}

@article{bai2024longalign,
  title={LongAlign: A Recipe for Long Context Alignment of Large Language Models},
  author={Bai, Yushi and Lv, Xin and Zhang, Jiajie and Lyu, Hongchang and Tang, Jiankai and Huang, Zhidian and Du, Zhengxiao and Liu, Xiao and Zeng, Aohan and Hou, Lei and Dong, Yuxiao and Tang, Jie and Li, Juanzi},
  journal={arXiv preprint arXiv:2401.18058},
  year={2024}
}

@article{jiang2024longllmlingua,
  title={LongLLMLingua: Accelerating and Enhancing LLMs in Long Context Scenarios via Prompt Compression},
  author={Jiang, Huiqiang and Wu, Qianhui and Luo, Chin-Yew and Li, Dongsheng and Lin, Chin-Yew and Yang, Yuqing and Qiu, Lili},
  journal={Proceedings of the 62nd Annual Meeting of the Association for Computational Linguistics},
  year={2024}
}

@article{wan2025qwenlongl1,
  title={QwenLong-L1: Towards Long-Context Large Reasoning Models with Reinforcement Learning},
  author={Wan, Fanqi and Shen, Weizhou and Liao, Shengyi and Shi, Yingcheng and Li, Chenliang and Yang, Ziyi and Zhang, Ji and Huang, Fei and Zhou, Jingren and Yan, Ming},
  journal={arXiv preprint arXiv:2505.17667},
  year={2025}
}

@inproceedings{jimenez2024swe,
  title={Swe-bench: Can language models resolve real-world github issues?},
  author={Jimenez, Carlos E and Yang, John and Wettig, Alexander and Yao, Shunyu and Pei, Kexin and Press, Ofir and Narasimhan, Karthik},
  booktitle={International Conference on Learning Representations},
  volume={2024},
  pages={54107--54157},
  year={2024}
}

@article{guo2025deepseek,
  title={Deepseek-r1: Incentivizing reasoning capability in llms via reinforcement learning},
  author={Guo, Daya and Yang, Dejian and Zhang, Haowei and Song, Junxiao and Wang, Peiyi and Zhu, Qihao and Xu, Runxin and Zhang, Ruoyu and Ma, Shirong and Bi, Xiao and others},
  journal={arXiv preprint arXiv:2501.12948},
  year={2025}
}

@article{team2026kimi,
  title={Kimi k3: Open frontier intelligence},
  author={Team, Kimi and Bai, Tongtong and Bai, Yifan and Bao, Yiping and Cai, Jianfeng and Cai, Xinyuan and Cao, Peizhou and Cao, Yuxuan and Chai, Ziwei and Charles, Y and others},
  journal={arXiv preprint arXiv:2607.24653},
  year={2026}
}

@misc{openai2025o3,
  title        = {OpenAI o3 and o4-mini System Card},
  author       = {{OpenAI}},
  year         = {2025},
  month        = apr,
  howpublished = {\url{https://openai.com/index/o3-o4-mini-system-card/}},
  note         = {Accessed 2026-09-25}
}

@article{hsieh2024ruler,
  title={RULER: What's the real context size of your long-context language models?},
  author={Hsieh, Cheng-Ping and Sun, Simeng and Kriman, Samuel and Acharya, Shantanu and Rekesh, Dima and Jia, Fei and Zhang, Yang and Ginsburg, Boris},
  journal={arXiv preprint arXiv:2404.06654},
  year={2024}
}

@article{liu2024lost,
  title={Lost in the middle: How language models use long contexts},
  author={Liu, Nelson F and Lin, Kevin and Hewitt, John and Paranjape, Ashwin and Bevilacqua, Michele and Petroni, Fabio and Liang, Percy},
  journal={Transactions of the association for computational linguistics},
  volume={12},
  pages={157--173},
  year={2024}
}

@article{robertson2009probabilistic,
  title={The Probabilistic Relevance Framework: BM25 and Beyond},
  author={Robertson, Stephen and Zaragoza, Hugo},
  journal={Foundations and Trends in Information Retrieval},
  volume={3},
  number={4},
  pages={333--389},
  year={2009}
}

@inproceedings{cormack2009reciprocal,
  title={Reciprocal Rank Fusion Outperforms Condorcet and Individual Rank Learning Methods},
  author={Cormack, Gordon V. and Clarke, Charles L. A. and Buettcher, Stefan},
  booktitle={Proceedings of the 32nd International ACM SIGIR Conference},
  pages={758--759},
  year={2009}
}

@article{shao2024deepseekmath,
  title={DeepSeekMath: Pushing the Limits of Mathematical Reasoning in Open Language Models},
  author={Shao, Zhihong and others},
  journal={arXiv preprint arXiv:2402.03300},
  year={2024}
}

@article{lewis2020retrieval,
  title={Retrieval-Augmented Generation for Knowledge-Intensive NLP Tasks},
  author={Lewis, Patrick and others},
  journal={Advances in Neural Information Processing Systems},
  volume={33},
  year={2020}
}

@article{zhang2024longbench,
  title={LongBench: A Bilingual, Multitask Benchmark for Long Context Understanding},
  author={Bai, Yushi and others},
  journal={Proceedings of ACL},
  year={2024}
}

@article{lin2004rouge,
  title={ROUGE: A Package for Automatic Evaluation of Summaries},
  author={Lin, Chin-Yew},
  journal={Text Summarization Branches Out},
  pages={74--81},
  year={2004}
}

@article{wang2025loongrl,
  title={LoongRL: Reinforcement Learning for Advanced Reasoning over Long Contexts},
  author={Wang, Siyuan and Zhang, Gaokai and Zhang, Li Lyna and Shang, Ning and Yang, Fan and Chen, Dongyao and Yang, Mao},
  journal={arXiv preprint arXiv:2510.19363},
  year={2025}
}

@article{chen2026longrlvr,
  title={LongRLVR: Long-Context Reinforcement Learning Requires Verifiable Context Rewards},
  author={Chen, Guanzheng and Shieh, Michael Qizhe and Bing, Lidong},
  journal={arXiv preprint arXiv:2603.02146},
  year={2026}
}

@article{lv2026golongrl,
  title={GoLongRL: Capability-Oriented Long Context Reinforcement Learning with Multitask Alignment},
  author={Lv, Minxuan and Mei, Tiehua and Du, Tanlong and Chen, Junmin and Su, Zhenpeng and Chen, Ziyang and Wang, Ziqi and Wu, Zhennan and Pan, Ruotong and Liang, Jian and Tang, Ruiming and Li, Han},
  journal={arXiv preprint arXiv:2605.19577},
  year={2026}
}

\clearpage
\appendix
\section{Supplementary Material}
\label{sec:appendix}
\renewcommand{\thefigure}{A\arabic{figure}}

This appendix provides additional experimental and implementation details, including training configurations, data statistics, process-level evaluation, component ablations, output-budget results, data-construction examples, and qualitative case studies.

\subsection*{AI use statement}
Generative AI tools were used solely to assist with language editing and polishing the presentation of this paper. All AI-assisted text was reviewed and revised by the authors, who take full responsibility for the final content.

\subsection{Detailed Experimental Setup}
\label{sec:appendix_setup}

\paragraph{Training hyperparameters.}
Both H2S models are trained with full-parameter updates from Qwen2.5-7B-Instruct-1M and Qwen2.5-14B-Instruct-1M. SFT runs for one epoch with a learning rate of $1\times10^{-5}$, a maximum input length of 128K, and a maximum target length of 4K. RL starts from the corresponding SFT checkpoint and uses full-parameter GRPO; each curriculum stage runs for one epoch with a learning rate of $1\times10^{-6}$, samples eight candidates per question, and allows at most 2K generated tokens. All stages use bf16, a cosine learning-rate schedule, 5\% warmup, gradient checkpointing, and ZeRO-3.

\paragraph{Reward implementation.}
The final-answer term follows each task's native output contract: numerical tolerance for numerical answers, choice accuracy for multiple-choice tasks, normalized exact match or token F1 for free-form QA, set F1 or NDCG for structured outputs, ROUGE-L for summarization, and joint content--citation quality for cited answers. The evidence term used during RL is span-level F1 over resolved source intervals; it is distinct from the block-level evidence recall used for the ESQ analysis in Section~\ref{sec:esq}. Citation coverage and information-need coverage are logged for diagnosis but are not additional terms in Equation~\ref{eq:h2s_reward}.

\paragraph{Comparison models.}
The main table compares five closed-source models, nine open-source models (including two Qwen2.5-Instruct base models), and our H2S-SFT and H2S variants at both model sizes. Every model is evaluated with a maximum input length of 128K and a maximum output length of 4K, and outputs that cannot be parsed receive a score of zero. We use each benchmark's native metric without additional model-specific prompt tuning.

\subsection{Training Data Overview}
\label{sec:appendix_data}

\begin{figure*}[!t]
\centering
\includegraphics[width=\textwidth]{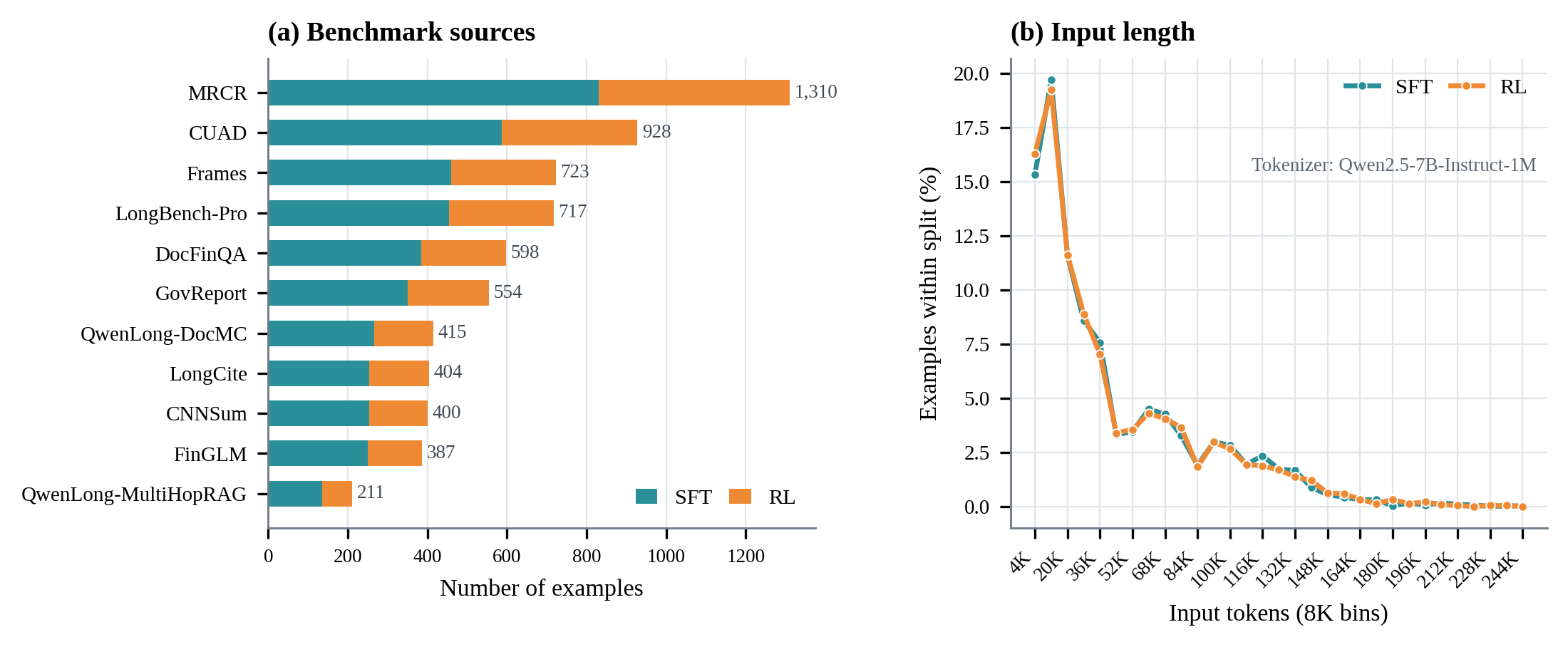}
\caption{\textbf{Benchmark and input-length distributions of the final training data.}
The left panel merges different task configurations from the same benchmark; LongBench-Pro contains T1--T11. The right panel shows the within-split length proportions for SFT and RL. The two splits have similar task-source and length distributions.}
\label{fig:appendix_data_distribution}
\end{figure*}

The final training set contains 4,228 SFT examples and 2,419 RL examples, for a total of 6,647 examples from 11 benchmark families, spanning precise retrieval, document question answering, reasoning, summarization, numerical calculation, and citation. Figure~\ref{fig:appendix_data_distribution} summarizes the benchmark sources and input-length distributions. The largest sources are MRCR, CUAD, Frames, and LongBench-Pro, while no single task accounts for most of the training set.

The SFT and RL splits have similar length profiles: approximately 55\% of examples are no longer than 32K tokens, while about 26\% exceed 64K tokens. The mean and median lengths are 44,097 and 28,290 tokens for SFT, and 43,656 and 27,726 tokens for RL, respectively. The 90th-percentile lengths are 108,497 tokens for SFT and 108,151 for RL. Examples longer than 128K are retained in the complete data resource but are not used in the current 128K training stages.

\subsection{Evidence--Summary Quality Subset and Detailed Results}
\label{sec:appendix_esq}

For case $i$, let $E_i$ denote recall over reference evidence blocks and $S_i$ denote summary ROUGE-L, both normalized to $[0,1]$. We define
\begin{equation}
\mathrm{ESQ}=\frac{100}{N}\sum_{i=1}^{N}\frac{2E_iS_i}{E_i+S_i},
\end{equation}
with the case-level value set to zero when both components are zero. The harmonic mean requires the model to perform well on both evidence selection and summary construction; a high score on either component cannot compensate for failure on the other.

\begin{table*}[t]
\centering
\scriptsize
\setlength{\tabcolsep}{4.2pt}
\renewcommand{\arraystretch}{1.06}
\caption{\textbf{Final-answer and evidence--summary quality.} Final Answer is the seven-task Avg from Table~\ref{tab:main_results}. E-Recall is the macro-average of per-example recall over reference evidence blocks; predictions are the 8-character block IDs parsed from \texttt{<evidence>}. Construction-time IDs are aligned to equivalent IDs in the actual inference blockization by normalized character matching before ID-only scoring. S-RL is ROUGE-L between the parsed \texttt{<summary>} and the reference summary. ESQ is the macro-average of the per-example harmonic mean of E-Recall and S-RL. Missing or unparsable components receive zero.}
\label{tab:esq_qc1090}
\resizebox{\textwidth}{!}{%
\begin{tabular}{llrrrr}
\toprule
Category & Model & Final Answer $\uparrow$ & E-Recall $\uparrow$ & S-RL $\uparrow$ & ESQ $\uparrow$ \\
\midrule
\multirow{5}{*}{\makecell[l]{Closed-source}}
& Claude Opus 5 & 43.13 & 56.31 & 23.93 & 25.45 \\
& GPT-5.5 & 42.65 & 41.10 & 28.61 & 21.87 \\
& DeepSeek-V4-Pro & 37.14 & 45.47 & 33.20 & 27.18 \\
& Kimi-K2.5 & 41.62 & 48.96 & 32.51 & 28.41 \\
& GLM-5.2 & 42.40 & 51.75 & 34.88 & 30.50 \\
\midrule
\multirow{7}{*}{Open-source}
& Qwen3-4B-Thinking-2507 & 14.59 & 14.94 & 11.70 & 6.44 \\
& Qwen3.8-27B & 22.43 & 2.71 & 3.01 & 1.18 \\
& Qwen3.5-35B-A3B & 9.52 & 15.32 & 7.61 & 6.73 \\
& QwenLong-L1-32B & 27.59 & 34.31 & 25.93 & 17.47 \\
& QwenLong-L1.5-30B-A3B & 10.76 & 7.74 & 5.30 & 4.17 \\
& Qwen2.5-7B-Instruct-1M & 9.55 & 14.49 & 18.16 & 5.80 \\
& Qwen2.5-14B-Instruct-1M & 19.14 & 28.32 & 25.84 & 15.61 \\
\midrule
\multirow{4}{*}{\textbf{Ours}}
& H2S-7B-SFT & 23.68 & 36.12 & 30.73 & 25.22 \\
& H2S-7B & 28.56 & 38.85 & 32.48 & 26.60 \\
& H2S-14B-SFT & 29.16 & 43.06 & 36.78 & 29.93 \\
& H2S-14B & 32.60 & 44.94 & 38.19 & 31.02 \\
\bottomrule
\end{tabular}}
\end{table*}

E-Recall is an ID-level measure rather than span or citation matching: each reference evidence block is counted once, and a predicted valid 8-character block UUID receives credit when it matches the construction-time ID or an equivalent ID located by normalized character matching in the actual inference input. This alignment is necessary because the stored construction annotations and inference inputs can use different block segmentations. The column is therefore reported as E-Recall, not Evidence F1. Summary quality is evaluated independently with ROUGE-L, and the ESQ column is the case-level harmonic mean of the two components.

\subsection{Detailed Output-Budget Results}
\label{sec:appendix_output_budget}

Table~\ref{tab:output_budget_full} provides the task-level results corresponding to Table~\ref{tab:budget}. Complete is the percentage of generations that either terminate before reaching the output limit or contain a complete final-answer field.

\begin{table*}[t]
\centering
\scriptsize
\setlength{\tabcolsep}{2.5pt}
\renewcommand{\arraystretch}{1.04}
\caption{\textbf{Detailed performance under different output budgets.}
Avg denotes the unweighted macro-average across the seven tasks, and Complete denotes the percentage of completed generations.}
\label{tab:output_budget_full}

\resizebox{\textwidth}{!}{%
\begin{tabular}{llrrrrrrrrr}
\toprule
Model & Budget & DocFinQA & Frames & LongCite & MRCR & AA-LCR & LongBenchV2 & HELMET-Summ & Avg & Complete \\
\midrule

\multirow{3}{*}{Qwen3-4B-Thinking-2507}
& 4K  & 14.61 & 18.71 & 3.68 & 25.45 & 9.88  & 18.99 & 10.80 & 14.59 & 53.75\% \\
& 8K  & 24.30 & 25.51 & 4.82 & 47.49 & 12.31 & 31.73 & 13.73 & 22.84 & 86.60\% \\
& 16K & 26.82 & 29.93 & 4.92 & 56.11 & 15.16 & 36.34 & 14.18 & 26.21 & 95.15\% \\
\midrule

\multirow{3}{*}{Qwen3.8-27B}
& 4K  & 40.52 & 32.31 & 4.94  & 46.89 & 15.69 & 15.14 & 1.51 & 22.43 & 41.17\% \\
& 8K  & 50.12 & 39.46 & 7.22  & 95.39 & 28.26 & 38.46 & 4.00 & 37.56 & 68.78\% \\
& 16K & 51.38 & 42.18 & 10.25 & 97.60 & 42.14 & 54.33 & 8.16 & 43.72 & 86.60\% \\
\midrule

\multirow{3}{*}{Qwen3.5-35B-A3B}
& 4K  & 24.72 & 18.03 & 2.13 & 8.62  & 6.11  & 6.97  & 0.06  & 9.52  & 14.83\% \\
& 8K  & 39.40 & 29.25 & 8.34 & 75.95 & 23.23 & 31.49 & 5.39  & 30.44 & 56.23\% \\
& 16K & 40.02 & 32.31 & 9.39 & 82.97 & 30.08 & 39.90 & 11.37 & 35.15 & 70.76\% \\
\midrule

\multirow{3}{*}{QwenLong-L1.5-30B-A3B}
& 4K  & 9.34  & 25.17 & 1.03 & 0.00 & 5.37  & 30.52 & 3.92  & 10.76 & 15.46\% \\
& 8K  & 23.83 & 36.73 & 2.00 & 1.40 & 27.03 & 33.65 & 5.67  & 18.62 & 53.86\% \\
& 16K & 30.72 & 40.82 & 5.23 & 6.61 & 36.45 & 48.32 & 10.78 & 25.56 & 86.91\% \\
\midrule

\multirow{3}{*}{H2S-7B}
& 4K  & 26.33 & 32.97 & 15.76 & 59.72 & 21.47 & 32.11 & 11.58 & 28.56 & 78.14\% \\
& 8K  & 26.33 & 33.33 & 17.16 & 61.93 & 22.21 & 35.77 & 16.02 & 30.39 & 92.00\% \\
& 16K & 26.33 & 33.33 & 17.17 & 62.16 & 22.21 & 35.77 & 16.42 & 30.48 & 93.13\% \\
\midrule

\multirow{3}{*}{H2S-14B}
& 4K  & 30.83 & 37.72 & 13.88 & 66.13 & 26.86 & 35.42 & 17.38 & 32.60 & 87.11\% \\
& 8K  & 30.83 & 37.72 & 14.54 & 67.24 & 27.41 & 35.97 & 20.40 & 33.44 & 93.05\% \\
& 16K & 30.83 & 37.72 & 14.54 & 67.44 & 27.41 & 35.97 & 20.98 & 33.56 & 93.98\% \\
\bottomrule
\end{tabular}%
}
\end{table*}

\subsection{LongBench-Pro Construction Example}
\label{sec:appendix_construction_case}

We use the RL example \texttt{v3\_lbp\_T10\_1249\_167a08600320} from LongBench-Pro to illustrate the complete transformation from a long document to a training target. The example contains 17,688 input tokens. Given a product manual, the question asks whether using oven cleaner on the non-stick coating of the AF-500 should be classified as \textit{Warning}, \textit{Caution}, or \textit{Note}.

\begin{enumerate}[leftmargin=*,label=\textbf{Step \arabic*.},itemsep=4pt]
\item \textbf{Block segmentation and retrieval.}
The document is segmented at semantic boundaries. BM25 and embedding retrieval are used in parallel, and the candidates are ranked with RRF. The two blocks directly relevant to the answer are:
\begin{quote}\small
\texttt{m6wyi3kt}: ``Do not use harsh chemicals (e.g., oven cleaner) on the non-stick surface---they will degrade the coating over time.''

\texttt{g7wu5mov}: ``Caution: Indicates a situation that may cause minor personal injury or slight appliance damage if not followed.''
\end{quote}

\item \textbf{Subquery decomposition and atomic-claim extraction.}
The question is decomposed into two information needs: the safety classification of using oven cleaner and the meaning of \textit{Caution} for the AF-500. The pipeline extracts two atomic facts: the operation is classified as \textit{Caution}, and \textit{Caution} indicates possible minor personal injury or slight appliance damage. A separate subquery about the CM-800 does not yield a supported claim and is excluded from the summary.

\item \textbf{Summary and answer construction.}
The supported claims are deduplicated and organized by information need to produce the following training target:
\begin{quote}\small
\textbf{Summary:} The AF-500 manual classifies using an oven cleaner on the non-stick coating as \textit{Caution}. In this manual, \textit{Caution} indicates possible minor personal injury or slight appliance damage if the instruction is not followed.

\textbf{Answer:} \texttt{[Answer] Caution}
\end{quote}

\item \textbf{Quality control.}
The pipeline checks block IDs, source-span alignment, claim--evidence alignment, question coverage, and whether the question can be answered using the summary alone. The evidence in this example matches the source text exactly, so the example is recorded with \texttt{stage6\_status=PASS} and \texttt{answerability=true}.
\end{enumerate}

This example shows that the pipeline does not construct a summary by directly concatenating retrieved blocks. It first removes unsupported or irrelevant information and then organizes traceable atomic facts into a question-conditioned working representation.

\subsection{LongBench-Pro Model Behavior}
\label{sec:appendix_model_behavior}

This section presents one successful case and one failure case to illustrate how structured training changes model behavior and where explicit compression can still fail. For readability, we retain only the evidence, summary, and final choice that determine the outcome.

\paragraph{Success case: inferring a missing option from stated information.}
The LongBench-Pro-T3 example \texttt{v3\_lbp\_T3\_0058\_ef8456ccf657} contains 8,720 input tokens. The question states that Yan'an, Suide, and Qingyang were among the regions where a unified progressive tax was piloted, and asks which of Guanzhong, Suide, Qingyang, and Yan'an was not mentioned. The key source text is:
\begin{quote}\small
\texttt{jidnkfv1}: ``In August 1943, a unified progressive tax was piloted in the three counties of Yan'an, Suide, and Qingyang.''
\end{quote}

Qwen2.5-14B-Instruct-1M selects D, while H2S-14B-SFT and H2S-14B both select the correct answer A. The trained models' intermediate outputs can be summarized as:
\begin{quote}\small
\textbf{Evidence:} The source lists only Yan'an, Suide, and Qingyang.

\textbf{Summary:} Among the four options, the region not listed in the source is Guanzhong.

\textbf{Answer:} \texttt{[Answer] A}
\end{quote}

The question is easy to localize but requires converting the three listed regions into the missing option. The explicit summary separates this relational judgment from final answer generation, allowing the model to answer directly from the organized representation.

\paragraph{Failure case: a claim extends beyond its evidence span.}
The LongBench-Pro-T11 example \texttt{v3\_lbp\_T11\_0602\_d12ef35fdad8} contains 31,340 input tokens and asks which statement about \textit{The Vegetarian} is strictly supported by the document; the correct answer is C. Qwen2.5-14B-Instruct-1M selects C, whereas H2S-14B-SFT and H2S-14B select B. The block identified by the model, \texttt{we85ooxe}, states that \textit{The Vegetarian} is a Booker International Prize-winning work by Han Kang, that vegetarianism is an extreme resistance to a patriarchal social system, and that Yeong-hye wishes to become plant-like without harming others. H2S-14B nevertheless produces the following claim:
\begin{quote}\small
``The novel was described by the International Booker Prize judges as unsettling, subtle, and beautiful, while exploring the body, desire, and social discipline.''
\end{quote}

The source does not contain this evaluation or attribute it to the judges. The unsupported claim is carried into the summary and causes the model to select B. The failure is therefore not a retrieval failure: the model found a relevant block, but the claim exceeded the cited span. This case shows that block-level provenance makes evidence traceable but does not guarantee claim faithfulness; future quality control should strengthen claim--span entailment checks, especially for entities, numbers, negation, and attribution.

\subsection{RL Length Curriculum}
\label{sec:appendix_curriculum}

We use a curriculum that progressively increases the context length:
\begin{equation}
\text{SFT}\ \longrightarrow\ \text{RL-32K}\ \longrightarrow\ \text{RL-64K}\ \longrightarrow\ \text{RL-128K}.
\end{equation}
The 32K stage is initialized from the SFT checkpoint, and each subsequent stage continues from the preceding checkpoint. The maximum sequence lengths for the 32K and 64K stages are 34,816 and 67,584, respectively. In the 128K stage, the 7B model uses 133,120 tokens and the 14B model uses 131,072 tokens. Each stage includes the prompt and at most 2,048 completion tokens.

Each stage runs for one epoch with a learning rate of $1\times10^{-6}$ and eight sampled candidates per question, using full-parameter GRPO, bf16, gradient checkpointing, and ZeRO-3. All stages use the same Evidence--Summary--Answer protocol and process rewards; only the permitted context length changes. The curriculum first stabilizes evidence selection and structured output on shorter documents, then gradually expands the range of information that the model must locate and compress.

\subsection{Question-Conditioned Summary Motivation}
\label{sec:appendix_summary_motivation}

Table~\ref{tab:summary_motivation_full} expands the DocQA-RL-1.6K motivating study in Section~\ref{sec:method}. Summaries are generated from only the question and document, without access to the reference answer. \textit{Claude Summary} denotes a summary generated by Claude Sonnet 4.6, while \textit{Qwen Self-Summary} denotes one generated by Qwen3.5-Flash itself in a separate pass. Context lengths are means computed uniformly with the Qwen2.5-7B-Instruct-1M tokenizer; a sum gives the document and added-summary lengths separately. Avg is the unweighted mean of the three document-length ranges.

\begin{table*}[t]
\centering
\scriptsize
\setlength{\tabcolsep}{3.6pt}
\renewcommand{\arraystretch}{1.08}
\caption{\textbf{Full results for question-conditioned summaries on long-document QA.} Accuracy is reported by document-length range and as their unweighted average; context tokens report the mean document and summary lengths, and $\Delta$ Avg is measured against LongDoc for the same answer model.}
\label{tab:summary_motivation_full}
\resizebox{\textwidth}{!}{%
\begin{tabular}{llrrrrrr}
\toprule
Answer model & Answer-time input & Context tokens & 0--20K & 20--40K & $>$40K & Avg & $\Delta$ Avg \\
\midrule
Claude Opus 4.6 & LongDoc & 15.19K & 98.9 & 89.9 & 78.3 & 89.0 & -- \\
Claude Opus 4.6 & Claude Summary & 0.46K & 100.0 & 96.6 & 95.7 & \textbf{97.4} & +8.4 \\
Claude Opus 4.6 & LongDoc + Claude Summary & 15.19K + 0.46K & 100.0 & 96.6 & 87.0 & 94.5 & +5.5 \\
\midrule
Qwen3.5-Flash & LongDoc & 15.19K & 77.3 & 75.3 & 65.2 & 72.6 & -- \\
Qwen3.5-Flash & Claude Summary & 0.46K & 97.7 & 94.4 & 87.0 & 93.0 & +20.4 \\
Qwen3.5-Flash & LongDoc + Claude Summary & 15.19K + 0.46K & 97.7 & 96.6 & 95.7 & \textbf{96.7} & +24.1 \\
Qwen3.5-Flash & LongDoc + Qwen Self-Summary (S) & 15.19K + 0.14K & 78.4 & 73.0 & 56.5 & 69.3 & -3.3 \\
Qwen3.5-Flash & LongDoc + Qwen Self-Summary (M) & 15.19K + 0.60K & 84.1 & 79.8 & 69.6 & 77.8 & +5.2 \\
Qwen3.5-Flash & LongDoc + Qwen Self-Summary (L) & 15.19K + 1.04K & 85.2 & 89.9 & 82.6 & 85.9 & +13.3 \\
\bottomrule
\end{tabular}}
\end{table*}

\begin{table*}[t]
\centering
\scriptsize
\setlength{\tabcolsep}{4pt}
\renewcommand{\arraystretch}{1.12}
\caption{\textbf{Construction statistics for the SFT and RL data.}
A case is an independent question--context instance; a block is the unit used for document segmentation and retrieval; a subquery is a decomposed information need; a claim is an independently verifiable atomic fact; and a citation links a claim to its source text. Quality-control counts follow their evaluated objects: faithfulness counts citations that exactly match the source, coverage counts subqueries supported by cited evidence, and answerability counts cases that can be solved from the summary alone. An additional 7,436 citations pass normalized fuzzy-0.95 matching and are reported in the appendix.}
\label{tab:data_construction}
\begin{tabular}{p{0.14\textwidth}p{0.19\textwidth}p{0.10\textwidth}rrrp{0.17\textwidth}}
\toprule
Stage & Statistic & Unit & SFT & RL & Total\\
\midrule
\multirow{3}{*}{\makecell[l]{Blocks \\ Retrieval}}
& Raw candidates & case & 5,300 & 3,000 & 8,300\\
& Segmentation & block & 1,127,610 & 657,020 & 1,784,630 \\
& Ranked retrieval & block & 193,031 & 109,550 & 302,581   \\
\addlinespace
\multirow{4}{*}{\makecell[l]{Claims \\Summaries}}
& Subquery decompose & subquery & 17,712 & 10,232 & 27,944  \\
& claim extraction & claim & 22,107 & 12,927 & 35,034 \\
& Canonical claims & claim & 21,211 & 12,325 & 33,536  \\
& Summary construct & case & 5,160 & 2,945 & 8,105 \\
\addlinespace
\multirow{4}{*}{\makecell[l]{Faithfulness \\ Coverage\\ Answerability}}
& Faithfulness & citation & 24,523 & 14,598 & 39,121 \\
& Coverage & subquery & 15,473 & 8,832 & 24,305\\
& Answerability & case & 4,539 & 2,600 & 7,139\\
& \textbf{Final data} & \textbf{case} & \textbf{4,228} & \textbf{2,419} & \textbf{6,647} \\
\bottomrule
\end{tabular}
\end{table*}

\subsection{Qualitative Case Studies}
\label{sec:appendix_detailed_cases}

We present eight curated case studies to illustrate how H2S transforms evidence
from long documents into question-conditioned summaries and final answers. The
cases cover numerical reasoning, cross-document synthesis, evidence-selection
errors, arithmetic errors, and evaluation sensitivity. Text shown under
\textbf{Evidence}, \textbf{Summary}, and \textbf{Answer} is reproduced verbatim
from the original model outputs. Long-document excerpts preserve the original
wording, with omitted content marked by \texttt{[...]}.

Among the eight examples, Cases~3 and~7 are labeled OOD under our evaluation
split. Case~3 illustrates successful cross-document evidence synthesis on
AA-LCR, whereas Case~7 illustrates a failure on LongBenchV2. These examples
provide qualitative views of H2S behavior under OOD evaluation; as curated
cases, they should not be interpreted as independent evidence of aggregate
OOD generalization.

\subsubsection{Successful Evidence Compression and Reasoning}

\paragraph{Case 1: Numerical reasoning from two evidence spans.}
\label{sec:case_docfinqa_success}
\begin{center}
\fontsize{8.2}{8.2}\selectfont
\begin{tcolorbox}[enhanced,breakable,colback=white,colframe=gray!75!black,
title={Case 1: Numerical Reasoning from Two Evidence Spans}]
\textbf{Split:} ID \quad \textbf{Input length:} 64,784 tokens \quad
\textbf{Models:} H2S-7B and H2S-14B

\medskip
\textbf{Original Question}

``what is the percentage change in annual operating cash flow from 2009 to 2010?''

\medskip
\textbf{Long Document Excerpt} \texttt{[vj2m69ie]}

\textbf{Liquidity and Capital Resources}

The following table presents selected financial information and statistics as
of and for the three years ended September 24, 2011 (in millions):

\begin{center}
\scriptsize
\begin{tabular}{lrrr}
\toprule
 & 2011 & 2010 & 2009 \\
\midrule
Cash, cash equivalents and marketable securities & \$81,570 & \$51,011 & \$33,992 \\
\texttt{[...]} & \texttt{[...]} & \texttt{[...]} & \texttt{[...]} \\
Annual operating cash flow & \$37,529 &
\textcolor{red!80}{\strut \$18,595} &
\textcolor{red!80}{\strut \$10,159} \\
\bottomrule
\end{tabular}
\end{center}

Cash, cash equivalents and marketable securities increased \$30.6 billion or
60\% during 2011. \texttt{[...]}

\medskip
\textbf{H2S-7B Output}

\textbf{Evidence}

\textbf{1)}
``Annual operating cash flow was \$18,595 million in 2010.''
\texttt{[vj2m69ie]}

\textbf{2)}
``Annual operating cash flow was \$10,159 million in 2009.''
\texttt{[vj2m69ie]}

\medskip
\textbf{Summary}

``Annual operating cash flow was \$10,159 million in 2009 and \$18,595
million in 2010. The percentage change from 2009 to 2010 is approximately
83.0\%.''

\medskip
\textbf{Answer}

``83\%''

\textbf{Stored score:} 1.0

\medskip
\textbf{H2S-14B Output}

\textbf{Evidence}

\textbf{1)}
``The annual operating cash flow for 2009 was \$10,159 million.''
\texttt{[vj2m69ie]}

\textbf{2)}
``The annual operating cash flow for 2010 was \$18,595 million.''
\texttt{[vj2m69ie]}

\medskip
\textbf{Summary}

``The annual operating cash flow was \$10,159 million in 2009 and \$18,595
million in 2010. Therefore, the percentage change from 2009 to 2010 is
(18,595 - 10,159) / 10,159 * 100 = 82.99\%.''

\medskip
\textbf{Answer}

``82.99\%''

\textbf{Stored score:} 1.0

\medskip
\textbf{Reference Answer}

``83\%''

\medskip
\textbf{Analysis}

The source block contains several financial variables and three years of
values. H2S selects only the two operands required by the question, assigns
them to the correct years, and explicitly performs the percentage-change
calculation in the summary. The final predictions are accepted under
numeric-tolerance evaluation.
\end{tcolorbox}
\captionof{figure}{\textbf{Numerical reasoning from a compact evidence set.}
Red highlighting marks \$18,595 under 2010 and \$10,159 under 2009 in the
source excerpt. The lower portion shows how these values are transformed into
calculation-ready summaries and final answers.}
\label{fig:case_docfinqa_success}
\end{center}

\paragraph{Case 2: Multi-hop reasoning across distributed passages.}
\label{sec:case_slush_puppie}
\begin{center}
\fontsize{8.2}{8.2}\selectfont
\begin{tcolorbox}[enhanced,breakable,colback=white,colframe=gray!75!black,
title={Case 2: Multi-Hop Reasoning across Distributed Passages}]
\textbf{Split:} ID \quad \textbf{Input length:} 2,059 tokens \quad
\textbf{Models:} H2S-7B and H2S-14B

\medskip
\textbf{Original Question}

"The flavored ice brand Slush Puppie is actually named after a food made from
cornmeal-batter. That food was itself originally named after a fish native to
South Carolina. What's the genus name of the fish?"

\medskip
\textbf{Long Document Excerpt 1} \texttt{[3d02d842]}

\textbf{History}

Slush Puppie founder Will Radcliff (1939--2014) decided to start the company
after seeing a slush-making machine at a Chicago trade fair in 1970. Radcliff,
his sister and their mother came up with the name
\textcolor{red!80!black}{"Slush Puppie" (based on hush puppy)} while sitting
on their front porch in Cincinnati, Ohio.

\medskip
\textbf{Long Document Excerpt 2} \texttt{[e6ef3604]}

\textbf{Hushpuppy}

A hush puppy (or hushpuppy) is a small, savory, deep-fried round ball made from
\textcolor{red!80!black}{cornmeal-based batter}. Hushpuppies are frequently served as a side dish with
seafood and other deep-fried foods.

\medskip
\textbf{Long Document Excerpt 3} \texttt{[8aaa80a7]}

\textbf{Name}

The first recorded use of the word "hush-puppy" dates to 1899. The name has
no verified origin. Etymology is attributed to a variety of sources. Although
there are many possible origins for the dish, it is found to be akin to
\textcolor{red!80!black}{"red-horse bread", named after the red horse fish of
the South Carolina rivers}. \texttt{[...]}

\medskip
\textbf{Long Document Excerpt 4} \texttt{[fc827d39]}

\textbf{Moxostoma}

\textcolor{red!80!black}{Moxostoma, the redhorses or jumprocks, is a genus of
North American ray-finned fish in the family Catostomidae}. \texttt{[...]}

\medskip
\textbf{H2S-7B Output}

\textbf{Evidence}

\textbf{1)}
"The flavored ice brand Slush Puppie is named after the food hushpuppy."
\texttt{[3d02d842]}

\textbf{2)}
"Hushpuppies are named after the red horse fish of the South Carolina rivers."
\texttt{[8aaa80a7]}

\textbf{3)}
"The genus name of the red horse fish is Moxostoma."
\texttt{[fc827d39]}

\textbf{Summary}

"The flavored ice brand Slush Puppie is named after the food hushpuppy.
Hushpuppies are named after the red horse fish of the South Carolina rivers.
The genus name of the red horse fish is Moxostoma."

\textbf{Answer}

"Therefore, the answer is Moxostoma."

\textbf{Stored score:} 1.0

\medskip
\textbf{H2S-14B Output}

\textbf{Evidence}

\textbf{1)}
"The flavored ice brand Slush Puppie is named after a food made from
cornmeal-batter, specifically hush puppies."
\texttt{[d7e9a65f, 3d02d842]}

\textbf{2)}
"Hush puppies are small, savory, deep-fried round balls made from cornmeal-based
batter." \texttt{[e6ef3604]}

\textbf{3)}
"The name hush puppy is attributed to a variety of sources, but it is found to
be akin to 'red-horse bread', named after the red horse fish of the South
Carolina rivers." \texttt{[8aaa80a7]}

\textbf{4)}
"The red horse fish is a member of the genus Moxostoma."
\texttt{[fc827d39]}

\textbf{Summary}

"The flavored ice brand Slush Puppie is named after a food made from
cornmeal-batter, specifically hush puppies. Hush puppies are small, savory,
deep-fried round balls made from cornmeal-based batter. The name hush puppy is
attributed to a variety of sources, but it is found to be akin to 'red-horse
bread', named after the red horse fish of the South Carolina rivers. The red
horse fish is a member of the genus Moxostoma. Therefore, the genus name of the
fish is Moxostoma."

\textbf{Answer}

"Therefore, the answer is Moxostoma."

\textbf{Stored score:} 1.0

\medskip
\textbf{Reference Answer}

"Therefore, the answer is Moxostoma."

\medskip
\textbf{Analysis}

"A readable multi-hop chain: Slush Puppie $\rightarrow$ hushpuppy
$\rightarrow$ red horse fish $\rightarrow$ genus Moxostoma. It clearly
demonstrates organization of scattered facts
before answering, and both models are correct."
\end{tcolorbox}
\captionof{figure}{\textbf{Multi-hop reasoning across distributed passages.}
The four source excerpts establish the Slush Puppie to hushpuppy to red horse
fish to Moxostoma chain, which both model summaries preserve.}
\label{fig:case_slush_puppie}
\end{center}

\paragraph{Case 3: OOD Cross-document synthesis over 123K tokens.}
\label{sec:case_aalcr_success}
\begin{center}
\fontsize{8.2}{8.2}\selectfont
\begin{tcolorbox}[enhanced,breakable,colback=white,colframe=gray!75!black,
title={Case 3: Cross-Document Synthesis over 123K Tokens}]
\textbf{Split:} OOD \quad \textbf{Input length:} 122,862 tokens \quad
\textbf{Models:} H2S-7B and H2S-14B

\medskip
\textbf{Original Question}

``Throughout all the documents, there are two main LLM threats that are
discussed recurrently. These issues are clearly stated in a paper where Baum is
one of the authors. In a different paper by Liu and Hu, these two threats are
categorised under what umbrella term?''

\medskip
\textbf{Long Document Excerpt 1} \texttt{[2nydst2j]}

\textbf{Soft Begging: Modular and Efficient Shielding of LLMs against
\textcolor{red!80!black}{Prompt Injection and Jailbreaking} based on Prompt Tuning}

Simon Ostermann, Kevin Baum, Christoph Endres, Julia Masloh, Patrick Schramowski

\textbf{Abstract}

\textcolor{red!80!black}{Prompt injection (both direct and indirect) and
jailbreaking} are now recognized as significant issues for large language
models (LLMs), particularly due to their potential for harm in
application-integrated contexts. This extended abstract explores a novel
approach to protecting LLMs from such attacks, termed ``soft begging.''
\texttt{[...]}

\medskip
\textbf{Long Document Excerpt 2} \texttt{[i88la43s]}

\textbf{Exploring Vulnerabilities and Protections in Large Language Models: A Survey}

Frank Weizhen Liu \quad Chenhui Hu

\textbf{Abstract}

As Large Language Models (LLMs) increasingly become key components in various
AI applications, understanding their security vulnerabilities and the
effectiveness of defense mechanisms is crucial. This survey examines the
security challenges of LLMs, focusing on two main areas:
\textcolor{red!80!black}{Prompt Hacking} and Adversarial Attacks, each with
specific types of threats. Under \textcolor{red!80!black}{Prompt Hacking},
we explore \textcolor{red!80!black}{Prompt Injection and Jailbreaking
Attacks}, discussing how they work, their potential impacts, and ways to
mitigate them. \texttt{[...]}

\medskip
\textbf{H2S-7B Output}

\textbf{Evidence}

\textbf{1)}
``The two main LLM threats discussed in the paper by Baum are prompt injection
and jailbreaking.'' \texttt{[2nydst2j]}

\textbf{2)}
``In the paper by Liu and Hu, these two threats are categorized under the
umbrella term 'Prompt Hacking'.'' \texttt{[i88la43s]}

\medskip
\textbf{Summary}

``The two main LLM threats discussed in the paper by Baum are prompt injection
and jailbreaking. In the paper by Liu and Hu, these two threats are categorized
under the umbrella term 'Prompt Hacking'.''

\medskip
\textbf{Answer}

``Therefore, the answer is Prompt Hacking.''

\textbf{Stored score:} 1.0

\medskip
\textbf{H2S-14B Output}

\textbf{Evidence}

\textbf{1)}
``The two main LLM threats discussed recurrently are prompt injection and
jailbreaking.'' \texttt{[i88la43s]}

\textbf{2)}
``In the Liu and Hu paper, these two threats are categorized under the
umbrella term 'Prompt Hacking'.'' \texttt{[i88la43s]}

\medskip
\textbf{Summary}

``The two main LLM threats discussed recurrently are prompt injection and
jailbreaking. In the Liu and Hu paper, these two threats are categorized under
the umbrella term 'Prompt Hacking'.''

\medskip
\textbf{Answer}

``Therefore, the answer is Prompt Hacking.''

\textbf{Stored score:} 1.0

\medskip
\textbf{Reference Answer}

``Prompt Hacking''

\medskip
\textbf{Analysis}

This example requires evidence from two different papers. The first block
identifies prompt injection and jailbreaking as the relevant threats, while the
second establishes the category relation between those threats and Prompt
Hacking. H2S preserves the provenance of both claims and combines them into a
directly answerable representation.
\end{tcolorbox}
\captionof{figure}{
\textbf{Successful OOD cross-document synthesis over a 122,862-token context.}
The two source excerpts are drawn from different papers. H2S identifies the
recurring threats and connects them to their umbrella category while
preserving source provenance.
}
\label{fig:case_aalcr_success}
\end{center}

\paragraph{Case 4: Ordinal retrieval and exact reproduction.}
\label{sec:case_mrcr_platforms}
\begin{center}
\fontsize{8.2}{8.2}\selectfont
\begin{tcolorbox}[enhanced,breakable,colback=white,colframe=gray!75!black,
title={Case 4: Ordinal Retrieval and Exact Reproduction}]
\textbf{Split:} ID \quad \textbf{Input length:} 64,331 tokens \quad
\textbf{Models:} H2S-7B and H2S-14B 

\medskip
\textbf{Original Question}

"Prepend cidZrU4URG to the 2nd (1 indexed) short essay about platforms. Do not
include any other text in your response."

\medskip
\textbf{Long Document Excerpt} \texttt{[f5jnuira]}

\textbf{User:} write a short essay about platforms

\textbf{Assistant:} \textcolor{red!80!black}{Platforms play a transformative role in our contemporary
society, serving as the backbone of digital interaction and connectivity.} In
the broadest sense, platforms are systems or technologies that facilitate
interactions, transactions, and collaborations between users, businesses, and
services. They have revolutionized the way we communicate, work, and engage
with the world around us, enabling an unprecedented level of accessibility and
efficiency.

A significant evolution in platforms has been seen with the rise of social
media, which has reshaped how we connect and share information. Social media
platforms like Facebook, Twitter, and Instagram have become integral parts of
daily life, providing spaces for individuals to express themselves, access news,
and engage with communities. \texttt{[...]}

In conclusion, platforms have become indispensable in today's digital
landscape, catalyzing unparalleled connectivity and innovation. They embody the
fusion of technology and human interaction, offering immense opportunities
while posing new challenges. \textcolor{red!80!black}{As platforms continue to evolve, it will be
paramount to address these challenges responsibly and harness their potential
to foster a more connected, equitable, and efficient world.}

\medskip
\textbf{H2S-7B Output}

\textbf{Evidence}

\textbf{1)}
"The second short essay about platforms is the one that begins with 'Platforms
play a transformative role in our contemporary society' and ends with 'foster a
more connected, equitable, and efficient world.'" \texttt{[f5jnuira]}

\textbf{2)}
"The final response is the string 'cidZrU4URG' prepended to the second short
essay about platforms." \texttt{[f5jnuira]}

\textbf{Summary}

"The second short essay about platforms is the one that begins with 'Platforms
play a transformative role in our contemporary society' and ends with 'foster a
more connected, equitable, and efficient world.' The final response is the
string 'cidZrU4URG' prepended to that essay."

\textbf{Answer}

"cidZrU4URGPlatforms play a transformative role in our contemporary society,
serving as the backbone of digital interaction and connectivity. In the
broadest sense, platforms are systems or technologies that facilitate
interactions, transactions, and collaborations between users, businesses, and
services. They have revolutionized the way we communicate, work, and engage
with the world around us, enabling an unprecedented level of accessibility and
efficiency.

\texttt{[...]}

In conclusion, platforms have become indispensable in today's digital
landscape, catalyzing unparalleled connectivity and innovation. They embody the
fusion of technology and human interaction, offering immense opportunities
while posing new challenges. As platforms continue to evolve, it will be
paramount to address these challenges responsibly and harness their potential
to foster a more connected, equitable, and efficient world."

\textbf{Stored score:} 1.0

\medskip
\textbf{H2S-14B Output}

\textbf{Evidence}

\textbf{1)}
"The second short essay about platforms is the one that begins with 'Platforms
play a transformative role in our contemporary society' and ends with 'a more
connected, equitable, and efficient world.'" \texttt{[f5jnuira]}

\textbf{Summary}

"The second short essay about platforms is the one that begins with 'Platforms
play a transformative role in our contemporary society' and ends with 'a more
connected, equitable, and efficient world.' Its full text is: 'Platforms play a
transformative role in our contemporary society, serving as the backbone of
digital interaction and connectivity. In the broadest sense, platforms are
systems or technologies that facilitate interactions, transactions, and
collaborations between users, businesses, and services. [...] As platforms
continue to evolve, it will be paramount to address these challenges
responsibly and harness their potential to foster a more connected, equitable,
and efficient world.'"

\textbf{Answer}

"cidZrU4URGPlatforms play a transformative role in our contemporary society,
serving as the backbone of digital interaction and connectivity. In the
broadest sense, platforms are systems or technologies that facilitate
interactions, transactions, and collaborations between users, businesses, and
services. They have revolutionized the way we communicate, work, and engage
with the world around us, enabling an unprecedented level of accessibility and
efficiency.

\texttt{[...]}

In conclusion, platforms have become indispensable in today's digital
landscape, catalyzing unparalleled connectivity and innovation. They embody the
fusion of technology and human interaction, offering immense opportunities
while posing new challenges. As platforms continue to evolve, it will be
paramount to address these challenges responsibly and harness their potential
to foster a more connected, equitable, and efficient world."

\textbf{Stored score:} 1.0

\medskip
\textbf{Reference Answer}

"cidZrU4URGPlatforms play a transformative role in our contemporary society,
serving as the backbone of digital interaction and connectivity. [...] As
platforms continue to evolve, it will be paramount to address these challenges
responsibly and harness their potential to foster a more connected, equitable,
and efficient world."

\medskip
\textbf{Analysis}

"A precise-retrieval case at about 64K tokens. Both models locate the second
essay, summarize its boundaries, and return the exact requested text with the
prefix."
\end{tcolorbox}
\captionof{figure}{\textbf{Ordinal retrieval and exact reproduction at about 64K tokens.}
The source excerpt contains the target essay, while the two trajectories
identify its ordinal position and reproduce it with the requested prefix.}
\label{fig:case_mrcr_platforms}
\end{center}

\subsubsection{Failure Modes and Evaluation Sensitivity}

\paragraph{Case 5: Evidence selection and arithmetic errors.}
\label{sec:case_docfinqa_failure}
\begin{center}
\fontsize{8.2}{8.2}\selectfont
\begin{tcolorbox}[enhanced,breakable,colback=white,colframe=gray!75!black,
title={Case 5: Evidence Selection and Arithmetic Errors}]
\textbf{Split:} ID \quad \textbf{Input length:} 100,115 tokens \quad
\textbf{Models:} H2S-7B and H2S-14B

\medskip
\textbf{Original Question}

``if you held 1000 shares on may 30 , 2014 , how much would you receive in
dividends?''

\medskip
\textbf{Distractor Document Excerpt} \texttt{[38eb3185]}

On February 10, 2015, we announced that our board of directors declared a
quarterly cash dividend on our common stock of
\textcolor{red!80}{\strut \$0.0675} per share. The dividend will be paid on
March 10, 2015 to all stockholders of record as of the close of business on
February 25, 2015.

\medskip
\textbf{Relevant Document Excerpt} \texttt{[9c9e1c33]}

\begin{center}
\scriptsize
\begin{tabular}{llll}
\toprule
Dividend Amount & Declaration Date & Record Date & Payment Date \\
\midrule
\$0.0425 & February 12, 2014 & February 25, 2014 & March 10, 2014 \\
\textcolor{red!80}{\strut \$0.0425} &
\textcolor{red!80}{\strut May 8, 2014} & May 27, 2014 & June 10, 2014 \\
\$0.0425 & July 31, 2014 & August 25, 2014 & September 10, 2014 \\
\$0.0675 & November 6, 2014 & November 25, 2014 & December 10, 2014 \\
\bottomrule
\end{tabular}
\end{center}

\medskip
\textbf{H2S-7B Output}

\textbf{Evidence}

\textbf{1)}
``On May 30, 2014, the company declared a quarterly
cash dividend of \$0.0675 per share.'' \texttt{[38eb3185]}

\textbf{2)}
``The dividend will be paid on March 10, 2015 to all stockholders of record as
of the close of business on February 25, 2015.''
\texttt{[38eb3185]}

\textbf{3)}
``If you held 1000 shares on May 30, 2014, you would
receive \$675 in dividends, calculated as 1000 shares multiplied by the
dividend per share of \$0.0675.'' \texttt{[38eb3185]}

\textbf{Summary}

``On May 30, 2014, the company declared a quarterly cash dividend of \$0.0675
per share. The dividend will be paid on March 10, 2015 to all stockholders of
record as of the close of business on February 25, 2015. If you held 1000
shares on May 30, 2014, you would receive \$675 in dividends, calculated as
1000 shares multiplied by the dividend per share of \$0.0675.''

\textbf{Answer}

``The correct answer is (B).''

\textbf{Stored score:} 0.0

\medskip
\textbf{H2S-14B Output}

\textbf{Evidence}

\textbf{1)}
``The dividend declared on May 8, 2014 was \$0.0425 per share.''
\texttt{[9c9e1c33]}

\textbf{2)}
``The dividend declared on May 8, 2014 was paid on June 10, 2014.''
\texttt{[9c9e1c33]}

\textbf{Summary}

``The dividend declared on May 8, 2014 was \$0.0425 per share, and it was paid
on June 10, 2014. Therefore, if you held 1000 shares on May 30, 2014, you would
receive \textcolor{red!75!black}{\$4.25} in dividends.''

\textbf{Answer}

``4.25''

\textbf{Stored score:} 0.0

\medskip
\textbf{Reference Answer}

``42.5''

\medskip
\textbf{Analysis}

The visible source blocks allow the two failures to be distinguished. H2S-7B
selects an unrelated 2015 dividend event and incorrectly attributes it to the
requested date. Its trajectory therefore fails during evidence selection and
subsequently introduces additional arithmetic and formatting errors. H2S-14B
selects the correct May 2014 dividend but computes $1{,}000\times0.0425$ as
$4.25$ rather than $42.5$, localizing its failure to downstream arithmetic.
\end{tcolorbox}
\captionof{figure}{\textbf{Different failure locations on the same question.}
The upper source portion highlights the incorrect 2015 block in red for
H2S-7B and the correct May 2014 table row in red for H2S-14B. The lower
portion compares the two Evidence--Summary--Answer trajectories.}
\label{fig:case_docfinqa_failure}
\end{center}

\paragraph{Case 6: Attribution and temporal-linking failure.}
\label{sec:case_gareau_failure}
\begin{center}
\fontsize{8.2}{8.2}\selectfont
\begin{tcolorbox}[enhanced,breakable,colback=white,colframe=gray!75!black,
title={Case 6: Attribution and Temporal-Linking Failure}]
\textbf{Split:} ID \quad \textbf{Input length:} 510 tokens \quad
\textbf{Models:} H2S-7B and H2S-14B 

\medskip
\textbf{Original Question}

"Rosie Ruiz was disqualified from the Boston Marathon. The subsequent winner
placed in what position of the 1988 Grandma's Marathon?"

\medskip
\textbf{Long Document Excerpt} \texttt{[a1d2b72b]}

\textbf{Jacqueline\_Gareau}

Jacqueline Gareau (born March 10, 1953) is a Canadian runner who
\textcolor{red!80!black}{won the Boston Marathon on April 21, 1980}. Gareau led
the women's field for most of the race,
only to find another runner, Rosie Ruiz, wearing the traditional victor's
laurels when she crossed the finish line. \textcolor{red!80!black}{Ruiz was
later disqualified after it was determined she had cheated and did not run the
entire race. Gareau was awarded the victory in a special ceremony one week
later.} Her official time for
the 1980 marathon, 2:34:28, was the fastest time recorded for a woman in the
event's history at the time.

Gareau met Ruiz two years after the marathon as she prepared to run a 10K race
in Miami, Florida. The encounter was brief and Ruiz refused (as she continued
to do until her death) to concede that she did not win the 1980 marathon.

Gareau served as the Grand Marshal of the 2005 Boston Marathon and was allowed
to "break the tape" in a special ceremony. She married her former coach,
Montreal banker Gilles Lapierre. She has a son, Yannick Lapierre, who
participates in Nordic skiing.

\medskip
\textbf{H2S-7B Output}

\textbf{Evidence}

\textbf{1)}
"Rosie Ruiz was disqualified from the Boston Marathon."
\texttt{[a1d2b72b]}

\textbf{2)}
"The subsequent winner of the Boston Marathon was Jacqueline Gareau, who placed
first." \texttt{[a1d2b72b]}

\textbf{3)}
"Jacqueline Gareau placed first in the 1988 Grandma's Marathon."
\texttt{[a1d2b72b]}

\textbf{Summary}

"Rosie Ruiz was disqualified from the Boston Marathon. The subsequent winner
of the Boston Marathon was Jacqueline Gareau, who placed first. Jacqueline
Gareau also placed first in the 1988 Grandma's Marathon."

\textbf{Answer}

"Therefore, the answer is First."

\textbf{Stored score:} 0.0

\medskip
\textbf{H2S-14B Output}

\textbf{Evidence}

\textbf{1)}
"Jacqueline Gareau won the Boston Marathon on April 21, 1980, after Rosie Ruiz
was disqualified." \texttt{[a1d2b72b]}

\textbf{2)}
"Jacqueline Gareau placed 10th in the 1988 Grandma's Marathon."
\texttt{[a1d2b72b]}

\textbf{Summary}

"Jacqueline Gareau won the Boston Marathon on April 21, 1980, after Rosie Ruiz
was disqualified. Gareau placed 10th in the 1988 Grandma's Marathon."

\textbf{Answer}

"Therefore, the answer is 10th."

\textbf{Stored score:} 0.0

\medskip
\textbf{Reference Answer}

"Therefore, the answer is Jacqueline Gareau placed first in the 1988 Grandma's
Marathon.."

\medskip
\textbf{Analysis}

"An attribution/temporal-linking failure: both models identify Jacqueline
Gareau as the subsequent Boston winner, but disagree on her 1988 Grandma's
Marathon placement and both score zero."
\end{tcolorbox}
\captionof{figure}{\textbf{Attribution and temporal-linking failure.} Both
trajectories identify Jacqueline Gareau as the subsequent Boston winner, but
their linked placement claims for the 1988 Grandma's Marathon disagree.}
\label{fig:case_gareau_failure}
\end{center}

\paragraph{Case 7: Evidence selection and summary drift.}
\label{sec:case_longbenchv2_failure}
\begin{center}
\fontsize{8.2}{8.2}\selectfont
\begin{tcolorbox}[enhanced,breakable,colback=white,colframe=gray!75!black,
title={Case 7: OOD Evidence Selection and Summary Drift}]
 \textbf{Split:} OOD \quad \textbf{Input length:} 15,515 tokens \quad
\textbf{Models:} H2S-7B and H2S-14B 

\medskip
\textbf{Original Question}

"What is the correct answer to this question: Which of the following
statements is incorrect?

Choices:

(A) By adjusting the residual gradually during the diffusion process, the model
can generate high-resolution images more efficiently.

(B) A complex noise control scheme was designed to flexibly control the
switching speed and noise intensity during the diffusion process.

(C) In the forward process, the optimization of $\theta$ is achieved by
minimizing the negative evidence lower bound

(D) The real data set consists of pictures taken by the camera, photos searched
on the Internet, and pictures used in literature"

\medskip
\textbf{Long Document Excerpt 1} \texttt{[fc428223]}

\textbf{Fig. 2.}

Construction process of training data. \textcolor{red!80!black}{Due to the lack
of publicly available sonar HR image dataset, we use natural image dataset for
training.} A large
degradation space is used to generate LR images. Moreover, we randomly sample
from the task-level LR to construct training-level LR.

\texttt{[...]}

Finally, we use the global residual connection to obtain the feature map, and
input it into the reconstruction module to generate the SR images
$S = \{s_1, s_2, \ldots\}$

$S = \mathrm{HRM}(Y_0 + Y_{\mathrm{DF}}) \qquad (9)$

\texttt{[...]}

\textcolor{red!80!black}{In training stage, we use L1 loss in pixelwise to
optimize network parameters}

$L = \sum \lVert s_i - x_i \rVert_1 \qquad (10)$

where b represents the task size of input and $\lVert\cdot\rVert_1$ denotes L1
norm.

\medskip
\textbf{Long Document Excerpt 2} \texttt{[097320dc]}

\textbf{IV. EXPERIMENT}

\textbf{A. Data}

\textcolor{red!80!black}{We adopt the DIV2K [51] as the training dataset, which
includes 800 training HR images.} As mentioned in Section III-A, we first construct a large-scale
degradation space to generate task-level LR, and then randomly sample to obtain
training-level LR. Finally, training HR--LR pairs are formed by HR and
training-level LR.

The testing datasets include synthetic datasets and real sonar image datasets.
For the synthetic datasets, we employ BSD100 [52], Urban100 [53], and General100
[54]. Each of these datasets has 100 HR images, and the simulated LR images are
generated by various degradation parameters. \textcolor{red!80!black}{For the
real sonar image datasets, we select three representative images from the
KLSG-II [55].} It should be noted
that these images do not have HR references.

\medskip
\textbf{H2S-7B Output}

\textbf{Evidence}

\textbf{1)}
"The real data set consists of pictures taken by the camera, photos searched on
the Internet, and pictures used in literature." \texttt{[097320dc]}

\textbf{2)}
"In the forward process, the optimization of $\theta$ is achieved by minimizing
the negative evidence lower bound." \texttt{[fc428223]}

\textbf{3)}
"A complex noise control scheme was designed to flexibly control the switching
speed and noise intensity during the diffusion process."
\texttt{[fc428223]}

\textbf{4)}
"By adjusting the residual gradually during the diffusion process, the model
can generate high-resolution images more efficiently."
\texttt{[fc428223]}

\textbf{Summary}

"The correct answer is (C). The real data set consists of pictures taken by
the camera, photos searched on the Internet, and pictures used in literature.
In the forward process, the optimization of $\theta$ is achieved by minimizing
the negative evidence lower bound. A complex noise control scheme was designed
to flexibly control the switching speed and noise intensity during the diffusion
process. By adjusting the residual gradually during the diffusion process, the
model can generate high-resolution images more efficiently."

\textbf{Answer}

"The correct answer is (C)."

\textbf{Stored score:} 1.0

\medskip
\textbf{H2S-14B Output}

\textbf{Evidence}

\textbf{1)}
"The training dataset used is the DIV2K dataset, which includes 800 training
HR images." \texttt{[097320dc]}

\textbf{2)}
"The real sonar image datasets include three representative images from the
KLSG-II dataset." \texttt{[097320dc]}

\textbf{3)}
"The training data construction process uses a natural image dataset for
training due to the lack of publicly available sonar HR image dataset."
\texttt{[fc428223]}

\textbf{Summary}

"The training dataset used is the DIV2K dataset, which includes 800 training
HR images. The real sonar image datasets include three representative images
from the KLSG-II dataset. The training data construction process uses a natural
image dataset for training due to the lack of publicly available sonar HR image
dataset. Therefore, the real data set does not consist of pictures taken by the
camera, photos searched on the Internet, and pictures used in literature; it
consists of natural images and sonar images from the KLSG-II dataset."

\textbf{Answer}

"The correct answer is (D)."

\textbf{Stored score:} 0.0

\medskip
\textbf{Reference Answer}

"C"

\medskip
\textbf{Analysis}

"A strict-support multiple-choice failure. H2S-7B is correct while H2S-14B
selects a different option after its summary focuses on irrelevant dataset
details, illustrating summary-induced distraction."
\end{tcolorbox}
\captionof{figure}{
\textbf{\textbf{Evidence selection and summary Edrift in an OOD example of H2S.}
H2S-7B identifies the incorrect statement as (C), whereas the H2S-14B
summary concentrates on dataset details and selects (D).
}}

\label{fig:case_longbenchv2_failure}
\end{center}

\end{document}